\documentclass{article} 
\usepackage{iclr2025_conference,times}

\usepackage{amsmath,amsfonts,bm}

\def\Figref#1{Figure~\ref{#1}}

\def\eqref#1{(\ref{#1})}

\def\1{\bm{1}}

\def\rvc{{\mathbf{c}}}

\def\rvp{{\mathbf{p}}}

\def\rvx{{\mathbf{x}}}
\def\rvy{{\mathbf{y}}}
\def\rvz{{\mathbf{z}}}

\DeclareMathAlphabet{\mathsfit}{\encodingdefault}{\sfdefault}{m}{sl}
\SetMathAlphabet{\mathsfit}{bold}{\encodingdefault}{\sfdefault}{bx}{n}

\def\gC{{\mathcal{C}}}

\def\gF{{\mathcal{F}}}

\def\gL{{\mathcal{L}}}

\def\gN{{\mathcal{N}}}

\def\gU{{\mathcal{U}}}

\newcommand{\E}{\mathbb{E}}

\newcommand{\R}{\mathbb{R}}

\usepackage{hyperref}
\hypersetup{backref=true,       
    pagebackref=true,               
    hyperindex=true,                
    colorlinks=true,                
    breaklinks=true,                
    urlcolor= orange,                
    linkcolor= orange,   
    bookmarks=true,                 
    bookmarksopen=false,
    filecolor=black,
    citecolor=blue,
    linkbordercolor=orange
}
\usepackage{url}
\usepackage{tabu}

\usepackage{amsthm,amssymb}
\usepackage{graphicx}
\usepackage[ruled,linesnumbered]{algorithm2e}
\usepackage{wasysym}
\usepackage{mathtools}
\usepackage{algorithmic}
\usepackage{subcaption}
\usepackage{wrapfig}
\usepackage{booktabs}
\usepackage{floatrow}

\theoremstyle{plain}
\newtheorem{theorem}{Theorem}[section]
\newtheorem{proposition}[theorem]{Proposition}
\newtheorem{lemma}[theorem]{Lemma}

\newtheorem{remark}[theorem]{Remark}

\newcommand{\deriv}{\mathrm{d}}
\usepackage{float}
\floatstyle{plaintop}
\restylefloat{table}
\usepackage{colortbl}
\usepackage{enumitem}
\usepackage{tikz}
\newcommand{\tikzxmark}{%
\tikz[scale=0.28] {
    \draw[line width=1,line cap=round] (0,0) to [bend left=6] (1,1);
    \draw[line width=1,line cap=round] (0.2,0.95) to [bend right=3] (0.8,0.05);
}}
\newcommand{\tikzcmark}{%
\tikz[scale=0.28] {
    \draw[line width=1,line cap=round] (0.25,0) to [bend left=10] (1,1);
    \draw[line width=1,line cap=round] (0,0.35) to [bend right=1] (0.23,0);
}}

\title{Semantic Image Inversion and Editing using Rectified Stochastic Differential Equations}

\author{
Litu Rout$^{1,2}$,\quad
Yujia Chen$^{1}$,\quad
Nataniel Ruiz$^{1}$,\\
\textbf{Constantine Caramanis}$^{2}$,\quad
\textbf{Sanjay Shakkottai}$^{2}$,\quad
\textbf{Wen-Sheng Chu}$^{1}$\\
$^1$ Google, $^2$ UT Austin\\
{\tt\small\{litu.rout,constantine,sanjay.shakkottai\}@utexas.edu}\\
{\tt\small\{yujiachen,natanielruiz,wschu\}@google.com}
}

\iclrfinalcopy 
\begin{document}

\maketitle
\vspace{-7ex}
\begin{figure*}[thb]
    \begin{center}
        \includegraphics[width=1\textwidth]{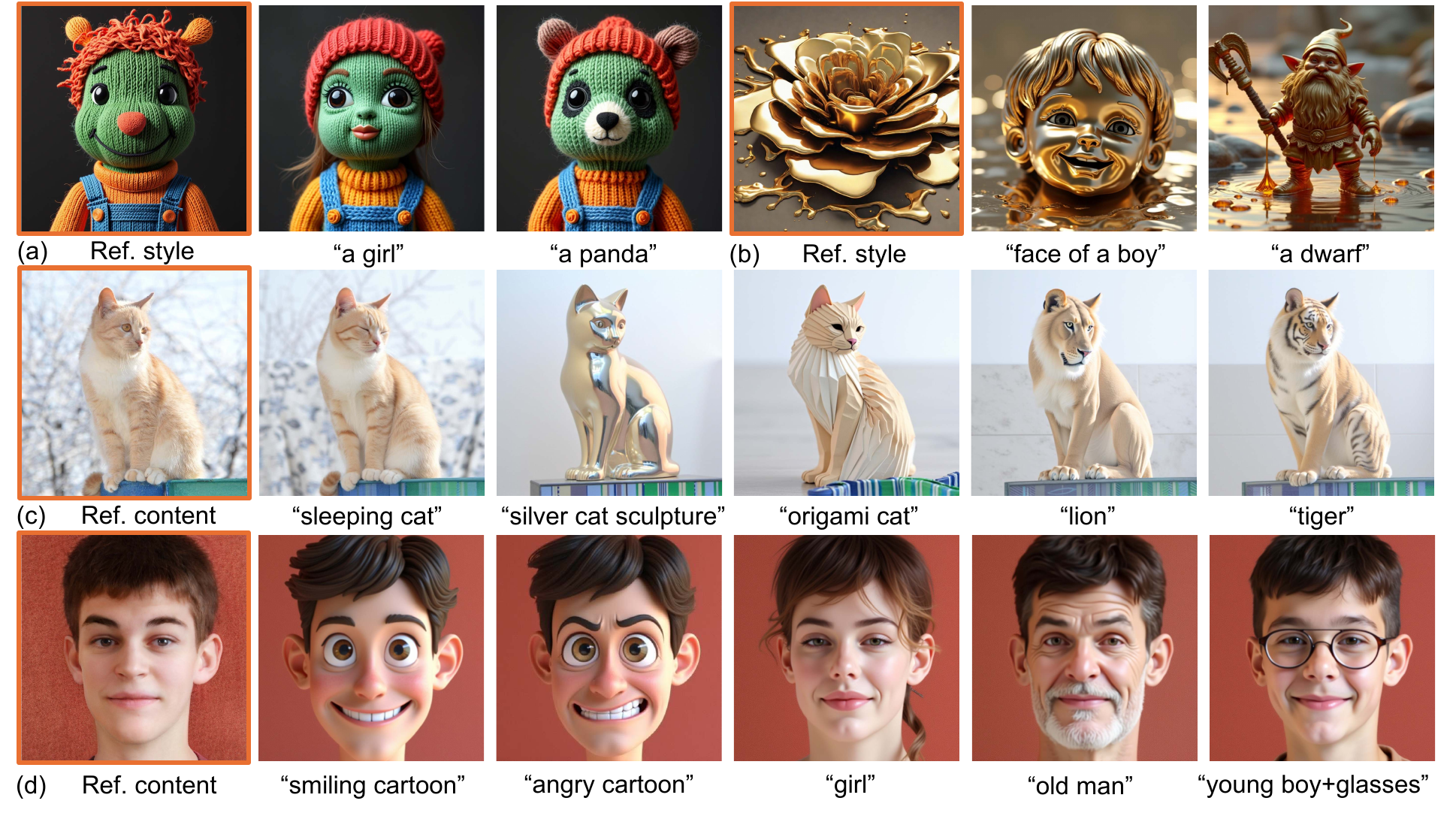}
    \end{center}
    \caption{
    \textbf{Rectified flows for image inversion and editing.}
    Our approach efficiently inverts reference style images in (a) and (b) without requiring text descriptions of the images and applies desired edits based on new prompts (e.g. ``a girl" or ``a dwarf").
   For a reference content image (e.g. a cat in (c) or a face in (d)), it performs semantic image editing (e.g. ``sleeping cat") and stylization (e.g. ``a photo of a cat in origmai style") based on prompts, without leaking unwanted content from the reference image. Input images have orange borders.
    }
    \label{fig:main}    
    \vspace{-3ex}    
\end{figure*}
\vspace{-1ex}

\begin{abstract}
\vspace{-2ex}
Generative models transform random noise into images; their inversion aims to transform images back to structured noise for recovery and editing.
This paper addresses two key tasks: (i) \textit{inversion} and (ii) \textit{editing} of a real image using stochastic equivalents of rectified flow models (such as Flux).
Although Diffusion Models (DMs) have recently dominated the field of generative modeling for images, their inversion presents faithfulness and editability challenges due to nonlinearities in drift and diffusion.
Existing state-of-the-art DM inversion approaches rely on training of additional parameters or test-time optimization of latent variables; both are expensive in practice.
Rectified Flows (RFs) offer a promising alternative to diffusion models, yet their inversion has been underexplored. 
We propose RF inversion using dynamic optimal control derived via a linear quadratic regulator. 
We prove that the resulting vector field is equivalent to a rectified stochastic differential equation. 
Additionally, we extend our framework to design a stochastic sampler for Flux.
Our inversion method allows for state-of-the-art performance in zero-shot inversion and editing, outperforming prior works 
in stroke-to-image synthesis and semantic image editing, with large-scale human evaluations confirming user preference.
\vspace{-2ex}
\end{abstract}

\section{Introduction}
\label{sec:intro}
\vspace{-1ex}

Vision generative models typically transform noise into images. 
Inverting such models, given a reference image, involves finding the structured noise that can regenerate the original image.
Efficient inversion must satisfy two crucial properties.
First, the structured noise should produce an image that is \textit{faithful} to the reference image.
Second, the resulting image should be easily \textit{editable} using new prompts, allowing fine modifications over the image.

Diffusion Models (DMs) have become the mainstream approach for generative modeling of images~\citep{sohl-dickstein15,ncsn,ddpm}, excelling at sampling from high-dimensional distributions~\citep{dalle, saharia2022palette, dalle2, ldm, sdxl, sc}.
The sampling process follows a Stochastic Differential Equation known as reverse SDE~\citep{anderson,efron2011tweedie,songscore}.
Notably, these models can invert a given image.
Recent advances in DM inversion have shown a significant impact on conditional sampling, such as stroke-to-image synthesis~\citep{sdedit}, image editing~\citep{p2p,nti,diffedit,rout2023theoretical,psld,stsl,indi} and stylization~\citep{stylealigned,rbm}.

Despite its widespread usage, DM inversion faces critical challenges in \textit{faithfulness} and \textit{editability}. 
First, the stochastic nature of the process requires fine discretization of the reverse SDE~\citep{ddpm, songscore}, which increases
expensive Neural Function Evaluations (NFEs). 
Coarse discretization, on the other hand, leads to less faithful outputs~\citep{sdedit}, even with deterministic methods like DDIM~\citep{ddim, songscore}.
Second, nonlinearities in the reverse trajectory introduce unwanted drift, reducing the accuracy of reconstruction~\citep{karras2024analyzing}.
While existing methods enhance faithfulness by optimizing latent variables~\citep{stsl} or prompt embeddings~\citep{nti,negprompt}, they tend to be less efficient, harder to edit, and rely on complex attention processors to align with a given prompt~\citep{p2p,stsl}.
These added complexities make such methods less suitable for real-world deployment.

\begin{figure}
\floatbox[{\capbeside\thisfloatsetup{capbesideposition={left,top},capbesidewidth=0.6\textwidth}}]{figure}[\FBwidth]
{\caption{
Graphical model illustrating (a) DDIM inversion and (b) RF inversion. 
Due to nonlinearities in DM trajectory, the DDIM inverted latent {\color{red}$\rvx_1$} significantly deviates from the original image {\color{orange} $\rvy_0$}. 
RF inversion without controller reduces this deviation, resulting in {\color{purple}$\rvx_1$}.
With controller, RF inversion further eliminates the reconstruction error, making {\color{cyan}$\rvx_1$} nearly identical to {\color{orange} $\rvy_0$}, which enhances the faithfulness.
}
\label{fig:graph}}
{\includegraphics[width=0.4\textwidth]{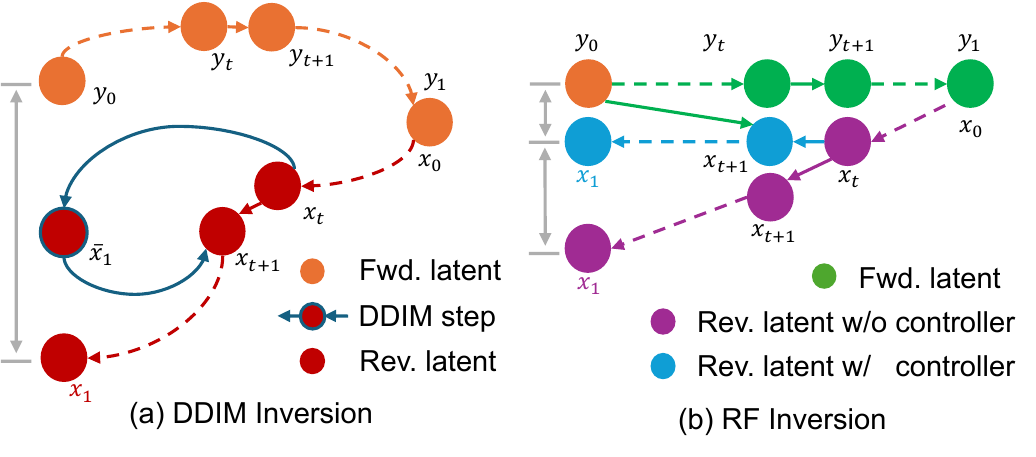}}
\vspace{-3ex}
\end{figure}

For inversion and editing, we introduce a zero-shot conditional sampling algorithm using Rectified Flows (RFs)~\citep{rectflow, interpolant, lipman2022flow, sd3}, a powerful alternative to DMs. 
Unlike DMs, where sampling is governed by a reverse SDE, RFs use an Ordinary Differential Equation known as reverse ODE, offering advantages in both efficient training and fast sampling. 
We construct a controlled forward ODE, initialized from a given image, to generate the initial conditions for the reverse ODE. 
The reverse ODE is then guided by an optimal controller, obtained through solving a Linear Quadratic Regulator (LQR) problem. 
We prove that the resulting new vector fields have a stochastic interpretation with an appropriate drift and diffusion. 
We evaluate RF inversion on stroke-to-image generation and image editing tasks, and show extensive qualitative results on other applications like cartoonization.
Our method significantly improves photo realism in stroke-to-image generation, surpassing a state-of-the-art (SoTA) method~\citep{nti} by 89\%, while maintaining faithfulness to the input stroke. 
In addition, we show that RF inversion outperforms DM inversion~\citep{sdedit} in faithfulness by 4.7\% and in realism by 13.8\% on LSUN-bedroom dataset~\citep{lsun}. 
\Figref{fig:main} and \Figref{fig:graph}  show the qualitative results of our approach and a graphical illustration, respectively.

Our theoretical and practical contributions can be summarized as:
\vspace{-1.5ex}
\begin{itemize}[left=3pt, noitemsep]
    \item We present an efficient inversion method for RF models, including Flux, that requires no additional training, latent  optimization, prompt tuning, or complex attention processors.
    
    \item We develop a new vector field for RF inversion, interpolating between two competing objectives: consistency with a possibly corrupted input image, and consistency with the ``true" distribution of clean images (\S\ref{sec:rf-inv}).
    We prove that this vector field is equivalent to a rectified SDE that interpolates between the stochastic equivalents of these competing objectives (\S\ref{sec:odes-vs-sdes}).
    We extend the theoretical results to design a stochastic sampler for Flux.
    
    \item We demonstrate the faithfulness and editability of RF inversion across three benchmarks: (i) LSUN-Bedroom, (ii) LSUN-Church, and (iii) SFHQ, on two tasks: stroke-to-image synthesis and image editing. 
    In addition, we provide extensive qualitative results and conduct large-scale human evaluations to assess user preference metrics (\S\ref{sec:exps}).
\end{itemize}

\vspace{-2.5ex}
\section{Related Works}
\label{sec:rel-work}
\vspace{-1.5ex}
\noindent\textbf{DM Inversion.} Diffusion models have become the mainstream approach for generative modeling, making DM inversion an exciting area of research~\citep{sdedit,diffedit,songscore,stylealigned,nti,stsl}. Among training-free methods, SDEdit~\citep{sdedit} adds noise to an image and uses the noisy latent as structured noise. For semantic image editing based on a given prompt, it simulates the standard reverse SDE starting from this structured noise. SDEdit requires no additional parameter training, latent variable optimization, or complex attention mechanisms. However, it is less faithful to the original image because adding noise in one step is equivalent to linear interpolation between the image and noise, while the standard reverse SDE follows a nonlinear path~\citep{rectflow,edm}.

An alternate method, DDIM inversion~\citep{ddim,songscore}, recursively adds predicted noise at each forward step and returns the final state as the structured noise (illustrated by $Y_t$ process in \Figref{fig:graph}(a)). 
However, DDIM inversion often deviates significantly from the original image due to nonlinearities in the drift and diffusion coefficients, as well as inexact score estimates~\citep{nti}.
To reduce this deviation, recent approaches optimize prompt embeddings~\citep{nti} or latent variables~\citep{stsl}, but they have high time complexity.
Negative prompt inversion~\citep{negprompt} speeds up the inversion process but sacrifices faithfulness. 
Methods like CycleDiffusion~\citep{cyclediffusion} and Direction Inversion~\citep{ju2023humansd} use inverted latents as references during editing, but they are either computationally expensive or not applicable to rectified flow models like Flux or SD3~\citep{sd3}.

\textbf{DM Editing.} 
Efficient inversion is crucial for real image editing. Once a structured noise is obtained by inverting the image, a new prompt is fed into the T2I generative model. 
Inefficient inversion often fails to preserve the original content and therefore requires complex editing algorithms. 
These editing algorithms can be broadly classified into (i) attention control, such as prompt-to-prompt~\citep{p2p}, plug-and-play (PnP)~\citep{pnp}, (ii) optimization-based methods like DiffusionCLIP~\citep{diffuseCLIP}, DiffuseIT~\citep{diffuseIT},  STSL~\citep{stsl}, and (iii)
latent masking to edit specific regions of an image using masks provided by the user~\citep{glide} or automatically extracted from the generative model~\citep{diffedit}. 
We focus on efficient inversion, avoiding the need for complex editing algorithms. 

\noindent\textbf{Challenges in RF Inversion.} 
Previous inversion or editing approaches have been tailored towards diffusion models and do not directly apply to SoTA rectified flow models like Flux. 
This limitation arises because the network architecture of Flux is MM-DiT~\citep{mmdit}, which is fundamentally different from the traditional UNet used in DMs~\citep{ddpm,ddim,songscore}.
In MM-DiT, text and image information are entangled within the architecture itself, whereas in UNet, text conditioning is handled via cross-attention layers. 
Additionally, Flux primarily uses T5 text encoder, which lacks an aligned latent space for images, unlike CLIP encoders.
Therefore, extending these prior methods to modern T2I generative models requires a thorough investigation. 
We take the first step by inverting and editing a given image using Flux.

\noindent\textbf{RF Inversion and Editing.}
DMs~\citep{ddpm,ddim,ldm} traditionally outperform RFs~\citep{lipman2022flow,rectflow,interpolant} in high-resolution image generation. 
However, recent advances have shown that RF models like Flux can surpass SoTA DMs in text-to-image (T2I) generation tasks~\citep{sd3}. 
Despite this, their inversion and editing capabilities remain underexplored.
In this paper, we introduce an efficient RF inversion method that avoids the need for training additional parameters~\citep{lora,dreambooth}, optimizing latent variables~\citep{stsl}, prompt tuning~\citep{nti}, or using complex attention processors~\citep{p2p}. 
While our focus is on inversion and editing, we also show that our framework can be easily extended to generative modeling.

\noindent\textbf{Filtering, Control and SDEs.}
There is a rich literature on the connections between nonlinear filtering, optimal control and SDEs~\citep{fleming2012deterministic,oksendal2003stochastic,tzen2019theoretical,pis}.
These connections are grounded in the Fokker-Planck equation~\citep{oksendal2003stochastic}, which RF methods~\citep{lipman2022flow,rectflow,interpolant,interpolants2} heavily exploit in sampling.
Our study focuses on rectified flows for conditional sampling, and shows that the resulting drift field also has an optimal control interpretation.

\vspace{-2ex}
\section{Method}
\label{sec:method}
\vspace{-1.5ex}
\subsection{Preliminaries}
\label{sec:prelim}
\vspace{-1.5ex}
In generative modeling, the goal is to sample from a target distribution $p_0$ given a finite number of samples from that distribution. 
Rectified flows~\citep{lipman2022flow,rectflow} represent a class of generative models that construct a source distribution $q_0$ and a time varying vector field $v_t(\rvx_t)$ to sample $p_0$ using an ODE:
\begin{align}
    \label{eq:gen-ode}
    \deriv X_t = v_t(X_t) \deriv t,\quad X_0 \sim q_0, \quad t\in [0,1].
\end{align}
Starting from $X_0=\rvx_0$, the ODE \eqref{eq:gen-ode} is integrated from $t:0\rightarrow 1$ to yield a sample $\rvx_1$ distributed according to $p_0$ (e.g., the distribution over images).
A common choice of $q_0$ is standard Gaussian $\gN\left(0,I\right)$ and $v_t\left(X_t\right) = -u(X_t,1-t;\varphi)$, where $u$ is a neural network parameterized by $\varphi$. The neural network is trained using the conditional flow matching objective as discussed below.

\noindent\textbf{Training Rectified Flows. } To train a neural network to serve as the vector field for the ODE~\eqref{eq:gen-ode}, we couple samples from $p_0$ with samples from $q_0$ -- which we call $p_1$ to simplify the notation -- via a linear path: $Y_t = t Y_1 + (1-t)Y_0$. The resulting marginal distribution of $Y_t$ becomes:
\begin{align}
    \label{eq:p-yt-marginal}
    p_t(\rvy_t) = \E_{Y_1 \sim p_1}\left[p_t(\rvy_t|Y_1) \right] = \int p_t(\rvy_t|\rvy_1) p_1(\rvy_1) \deriv\rvy_1.
\end{align}
Given an initial state $Y_0 = \rvy_0$ and a terminal state $Y_1=\rvy_1$, the linear path induces an ODE: $\deriv Y_t = u_t\left(Y_t|\rvy_1\right) \deriv t$ with the conditional vector field $u_t\left(Y_t|\rvy_1\right) = \rvy_1 - \rvy_0$.
The marginal vector field is derived from the conditional vector field using the following relation~\citep{lipman2022flow}:
\begin{align}
    \label{eq:ut-marginal}
    u_t(\rvy_t) = \E_{Y_1 \sim p_1}\left[u_t\left(\rvy_t|Y_1\right) \frac{p_t(\rvy_t|Y_1)}{p_t(\rvy_t)} \right]
    = \int u_t\left(\rvy_t|\rvy_1\right) \frac{p_t(\rvy_t|\rvy_1)}{p_t(\rvy_t)} p_1(\rvy_1) \deriv \rvy_1.
\end{align}
We can then use a neural network $u(\rvy_t,t;\varphi)$, parameterized by $\varphi$, to approximate the marginal vector field $u_t(\rvy_t)$ through the flow matching objective defined as:
\begin{align}
    \label{eq:fm-loss}
    \gL_{FM}(\varphi)\coloneqq \E_{t\sim \gU[0,1],Y_t\sim p_t}\left[\left\| u_t(Y_t) - u(Y_t,t;\varphi) \right\|_2^2\right].
\end{align}
For tractability, we can instead consider a different objective, called conditional flow matching:
\begin{align}
    \label{eq:cfm-loss}
    \gL_{CFM}(\varphi)\coloneqq \E_{t\sim \gU[0,1],Y_t\sim p_t(\cdot|Y_1), Y_1 \sim p_1}\left[\left\| u_t(Y_t|Y_1) - u(Y_t,t;\varphi) \right\|_2^2\right].
\end{align}
$\gL_{CFM}$ and $\gL_{FM}$ have the identical gradients~(\citealp[Theorem 2]{lipman2022flow}), and are hence equivalent.
However, $\gL_{CFM}(\varphi)$ is computationally tractable, unlike $\gL_{FM}(\varphi)$, and therefore preferred during training. 
Finally, the required vector field in \eqref{eq:gen-ode} is computed as $v_t\left(X_t\right) = -u(X_t,1-t;\varphi)$. 
In this way, rectified flows sample a data distribution by an ODE with a learned vector field.

\subsection{Connection between Rectified Flows and Linear Quadratic Regulator}
\label{sec:rf-lqr}
The unconditional rectified flows (RFs) (e.g., Flux) from Section \S\ref{sec:prelim} above, enable image generation by simulating the vector field $v_t(\cdot)$ initialized with a sample of random noise. Subsequently, by simulating the reversed vector field $-v_{1-t}(\cdot)$ starting from the image, we get back the sample of noise that we started with.  We formalize this statement below.
\begin{proposition}
\label{prop:inv-wo-control}
Given an image $\rvy_0$ and the vector field $v_t(\cdot)$ of the generative ODE~\eqref{eq:gen-ode}, suppose the structured noise $\rvy_1$ is obtained by simulating an ODE:
\begin{align}
\label{eq:inv-wo-control}
    \deriv Y_t = u_t(Y_t) \deriv t, \quad Y_0 = \rvy_0, \quad t\in[0,1]. 
\end{align} 
If $u_t(\cdot) = -v_{1-t}(\cdot)$ and $X_0 = \rvy_1$, then the ODE~\eqref{eq:gen-ode} recovers the original image, i.e., $X_1 = \rvy_0$.
\end{proposition}

\noindent\textbf{Implication.} Rectified flows enable exact inversion of a given image when the vector field of the generative ODE~\eqref{eq:gen-ode} is precisely known. 
Employing ODE~\eqref{eq:inv-wo-control} for the structured noise and ODE~\eqref{eq:gen-ode} to transform that noise back into an image, RF inversion accurately recovers the given image.

Suppose instead that we start with a {\em corrupted image} and simulate the reversed vector field $-v_{1-t}(\cdot)$. Then we obtain a noise sample. There are two salient aspects of this noise sample. First, it is consistent with the original image: when processed through $v_{t}(\cdot)$ it results in the same corrupted image. Second, if the image sample is ``atypical'' (e.g., corrupted, or, say, a stroke painting as in \S\ref{sec:exps}), then the sample of noise is also likely to be atypical. In other words, the noise sample is only consistent to the (possibly corrupted) image sample.

Our goal is to modify the pipeline above so that even when we start with a corrupted image, we can get back a clean image  (see stroke-to-image synthesis in ~\Figref{fig:stroke2image}), but for this, we need to processs by $v_{t}(\cdot)$  a noise sample that is closer to being ``typical''.
More generally, the goal is to create a pipeline that supports semantic editing of real images (\S\ref{sec:exps}), e.g., changing age, or gender without relying on additional training, optimization, or complex attention processors.

Thus, as a first step, we derive an optimal controller that takes a minimum energy path to convert any image $Y_0$ (whether corrupted or not) to a given sample of random noise $Y_1 \sim p_1$ -- i.e., noise that is typical for $p_1$.
Specifically, we consider optimal control 
in a $d$-dimensional vector space $\R^d$:
\begin{align}
    \label{eq:gen-lqr}
    V(c) \coloneqq  \int_0^1 \frac{1}{2}\left\|c\left(Z_t, t\right)\right\|_2^2 \deriv t + \frac{\lambda}{2} \left\|Z_1 - Y_1\right\|_2^2, ~\deriv Z_t = c\left(Z_t,t\right) \deriv t, ~Z_0 =\rvy_0,~Y_1 \sim p_1,
\end{align}
where $\lambda$ is the weight assigned to the terminal cost and $V(c)$ denotes the total cost of the control $c: \R^d\times [0,1] \rightarrow \R^d$. 
The minimization of $V(c)$ over the admissible set of controls, denoted by $\gC$, is known as the Linear Quadratic Regulator (LQR) problem.
The solution of the LQR problem \eqref{eq:gen-lqr} is given in \textbf{Proposition~\ref{prop:rf-lqr}}, which minimizes the quadratic transport cost of the dynamical system. 
\begin{proposition}
\label{prop:rf-lqr}
For $Z_0 =\rvy_0$ and $Y_1 = \rvy_1$, the optimal controller of the LQR problem \eqref{eq:gen-lqr}, denoted by $c^*\left(\cdot,t\right)$ is equal to the conditional vector field $u_t\left(\cdot|\rvy_1\right)$ of the rectified linear path $Y_t = tY_1 + (1-t)Y_0$ when $Y_0 = \rvy_0$, i.e., $c^*\left(\rvz_t,t\right) = u_t\left(\rvz_t|\rvy_1\right) = (\rvy_1 - \rvz_t)/(1-t)$.
\end{proposition}


\subsection{Inverting Rectified Flows With Dynamic Control}
\label{sec:rf-inv}
So far, we have two vector fields. The first, from the RFs, transforms an image $Y_0$ typical for distribution $p_0$ to a typical sample of random Gaussian noise $Y_1 \sim p_1.$ As discussed above, if the image sample is atypical, then the sample of noise is also likely to be atypical. 

We also have a second vector field resulting from the optimal control formulation that transforms {\em any} image (whether corrupted or not) to a noise sample that is typical-by-design from the distribution $p_1$. Therefore, this sample, when passed through the rectified flow ODE~\eqref{eq:gen-ode} results in a ``typical'' image from the ``true'' distribution $p_0$. This image is clean, i.e., typical for $p_0$, but it is not related to the image $Y_0$.
Our controlled ODE, defined below, interpolates between these two differing objectives -- {\em consistency with the given (possibly corrupted) image, and consistency with the distribution of images $p_0$} -- with a tunable parameter $\gamma$:
\begin{align}
    \label{eq:controlled-ODE}
    \deriv Y_t = \Big[u_t(Y_t) + \gamma \left(u_t(Y_t|\rvy_1) - u_t(Y_t)\right) \Big] \deriv t, \quad Y_0 = \rvy_0,
\end{align}
where $u_t(Y_t|\rvy_1) = c^*(Y_t,t)$ is computed based on the insights from \textbf{Proposition~\ref{prop:rf-lqr}}, and $u_t(Y_t) = -v_{1-t}(Y_t)$ as established in \textbf{Proposition~\ref{prop:inv-wo-control}}.
Here, we call $\gamma \in [0,1]$ the {\em controller guidance}. 
Thus, ODE~\eqref{eq:controlled-ODE} generalizes \eqref{eq:inv-wo-control} to editing applications, while keeping its inversion accuracy comparable.

When $\gamma=1$, the drift field of the ODE~\eqref{eq:controlled-ODE} becomes optimal controller of LQR problem~\eqref{eq:gen-lqr}, ensuring that the structured noise $Y_1 = \rvy_1$ adheres to the distribution $p_1$. Consequently, initializing the generative ODE~\eqref{eq:gen-ode} with $\rvy_1$ results in samples with high likelihood under the data distribution $p_0$. 

Conversely, when $\gamma = 0$, the system follows the ODE~\eqref{eq:inv-wo-control} described in \textbf{Proposition~\ref{prop:inv-wo-control}}, resulting a structured noise $Y_1$ that is not guaranteed to follow the noise distribution $p_1$.
However, initializing the generative ODE~\eqref{eq:gen-ode} with this noise precisely recovers the reference image $\rvy_0$.

Beyond this vector field interpolation intuition, we show in the next section \S\ref{sec:odes-vs-sdes} that the controlled ODE~\eqref{eq:controlled-ODE} has an SDE interpretation. As is well known \citep{ddpm,ddim,sdedit,songscore}, SDEs are robust to initial conditions, in proportion to the variance of the additive noise. Specifically, errors propagate over time in an ODE initialized with an incorrect or corrupted sample. However, SDEs (Markov processes) under appropriate conditions converge to samples from a carefully constructed invariant distribution with reduced sensitivity to the initial condition, resulting in a form of robustness to initialization. As we see, the parameter $\gamma$ (the controller guidance) appears in the noise term to the SDE, thus the SDE analysis in the next section again provides intuition on the trade-off between consistency to the (corrupted) image and consistency to the terminal invariant distribution.

\begin{remark}
\label{rmk:controlled-ode}
We note that our analysis extends to the case where $\gamma$ is time-varying, though we omit these results for simplicity of notation. 
This is useful in practice, especially when $\rvy_0$ is a corrupted image, because for large $\gamma$ the stochastic evolution~\eqref{eq:stoch-rect-flow-sampling} moves toward a sample from the invariant measure $\gN\left(0,I \right)$.
This noise encodes clean images. 
Starting from this noise, the corresponding reverse process operates in pure diffusion mode, resulting in a clean image.
As the process approaches the terminal state, $\gamma$ is gradually reduced to ensure that $\rvy_0$ is encoded through $u_t(\cdot)$ into the final structured noise sample.
\end{remark}

\subsection{Controlled Rectified Flows as Stochastic Differential Equations}
\label{sec:odes-vs-sdes}
An SDE~\citep{ddpm} is known to have an equivalent ODE formulation~\citep{ddim} under certain regularity conditions~\citep{anderson,songscore}.
In this section, we derive the opposite: an SDE formulation for our controlled ODE~\eqref{eq:controlled-ODE}  from \S\ref{sec:rf-inv}. 
Let $W_t$ be a $d$-dimensional Brownian motion in a filtered probability space $(\Omega, \gF, \{\gF_t\}, \mathbb{P})$.

\begin{theorem}
\label{theorem:sde-equiv-cntrl-ode}
Fix any $T \in (0, 1)$. For any $t \in [0, T]$, the controlled ODE~\eqref{eq:controlled-ODE} is explicitly given by:
\begin{align}
\label{eq:ode-controlled-vector-field}
    \deriv Y_t = \left[- \frac{1}{1-t}\left(Y_t - \gamma  \rvy_1\right) - \frac{(1-\gamma)t}{1-t} \nabla \log p_t(Y_t)\right]\deriv t,\quad Y_0 \sim p_0.
\end{align}
Its density evolution is identical to the density evolution of the following SDE:
\begin{align}
    \label{eq:sde-controlled-vector-field}
    \deriv Y_t = -\frac{1}{1-t}\left( Y_t -\gamma  \rvy_1\right) \deriv t + \sqrt{\frac{2(1-\gamma)t}{1-t}} \deriv W_t, \quad Y_0\sim p_0.
\end{align}
Finally, denoting $p_t(\cdot)$ as the marginal pdf of $Y_t$, the density evolution is explicitly given by: 
\begin{align}
    \frac{\partial p_t(Y_t)}{\partial t} = \nabla \cdot \left[\left( \frac{1}{1-t}\left(Y_t - \gamma  \rvy_1\right) + \frac{(1-\gamma)t}{1-t} \nabla \log p_t(Y_t)\right)p_t(Y_t)\right].
\end{align}
\end{theorem}

\textbf{Properties of SDE~\eqref{eq:sde-controlled-vector-field}.}
Elaborating on the intuition discussed at the end of \S\ref{sec:rf-inv}, 
when the controller guidance parameter $\gamma=0$, it becomes the stochastic equivalent of the standard RFs; see \textbf{Lemma~\ref{lemma:sde2ode}} for a precise statement. 
The resulting SDE is given by 
\begin{align}
    \label{eq:sde-optimal-vector-field}
    \deriv Y_t = -\frac{1}{1-t}Y_t \deriv t + \sqrt{\frac{2t}{1-t}} \deriv W_t, \quad Y_0\sim p_0,
\end{align}
which improves faithfulness to the image $Y_0$.
When $\gamma=1$, the SDE~\eqref{eq:sde-controlled-vector-field} solves the LQR problem \eqref{eq:gen-lqr} and drives towards the terminal state $Y_1 = \rvy_1$. This improves the generation quality, because the sample $Y_1$ is from the correct noise distribution $p_1$ as previously discussed in \S\ref{sec:rf-inv}. 
Therefore, a suitable choice of $\gamma$ retains faithfulness while simultaneously applying the desired edits.

Finally, we assume $T=1-\delta$ for sufficiently small $\delta$ (such that $0<\delta \ll 1$) to avoid irregularities at the boundary. This is typically considered in practice for numerical stability (even for diffusion models). Thus, in practice, the final sample $\rvy_{1-\delta}$ is returned as $\rvy_1$.

\noindent\textbf{Comparison with DMs.} 
Analogous to the SDE~\eqref{eq:sde-optimal-vector-field}, the stochastic noising process of DMs is typically modeled by the Ornstein-Uhlenbeck (OU) process, governed by the following SDE:
\begin{align}
    \label{eq:ddpm}
    \deriv Y_t = -Y_t \deriv t + \sqrt{2} \deriv W_t.
\end{align}
The corresponding ODE formulation is given by:
\begin{align}
    \label{eq:ddim}
    \deriv Y_t = \left[-Y_t - \nabla \log p_t(Y_t)\right] \deriv t .
\end{align}
Instead, our approach is based on rectified flows~(\ref{eq:gen-ode}), which leads to a different ODE and consequently translates into a different SDE. 
As an additional result, we formalize the ODE derivation in \textbf{Lemma~\ref{lemma:optimal-vector-field}}.
In \textbf{Lemma~\ref{lemma:sde2ode}}, we show that the marginal distribution of this ODE is equal to that of an SDE with appropriate drift and diffusion terms.
In \textbf{Proposition~\ref{prop:rect-flow-ode-equals-sde}}, we show that the stationary distribution of this new   SDE~\eqref{eq:sde-optimal-vector-field} converges to the standard Gaussian $\gN(0,I)$ in the limit as $t\rightarrow 1$.

The standard OU process~\eqref{eq:ddpm} interpolates between the data distribution at time $t=0$ and a standard Gaussian as $t \to \infty$. The SDE \eqref{eq:sde-optimal-vector-field}, however, interpolates between the data distribution at time $t=0$ and a standard Gaussian at $t = 1$. In other words, it effectively ``accelerates'' time as it progresses to achieve the terminal Gaussian distribution. This is accomplished by modifying the coefficients of drift and diffusion as in \eqref{eq:sde-optimal-vector-field} to depend explicitly on time $t$. Thus, a sample path of \eqref{eq:sde-optimal-vector-field} appears like a noisy line, unlike that of the OU process (see Appendix~\ref{sec:num_sim} for numerical simulations).

\vspace{-1ex}
\subsection{Controlled Reverse Flow using Rectified ODEs and SDEs}
\label{sec:controlled-reverse-ode-sde}
\vspace{-1ex}
In this section, we develop an ODE and an SDE similar to our discussions above, but for the reverse direction (i.e., from noise to images).

\textbf{Reverse process using ODE.} Starting from the structured noise $\rvy_1$ obtained by integrating the controlled ODE~\eqref{eq:controlled-ODE}, we construct another controlled ODE~\eqref{eq:gen-ode-w-controller} for the reverse process (i.e., noise to image).
In this process, the optimal controller uses the reference image $\rvy_0$ for guidance:
\begin{align}
    \label{eq:gen-ode-w-controller}
    \deriv X_t = \Big[v_t(X_t) + \eta \left( v_t(X_t|\rvy_0) - v_t(X_t) \right)\Big] \deriv t, \quad X_0 = \rvy_1, \quad t\in [0,1],
\end{align}
where $\eta\!\in\![0,1]$ is the {\em controller guidance parameter} as before that controls faithfulness and editability of the given image $\rvy_0$.
Similar to the analysis in 
\textbf{Proposition~\ref{prop:rf-lqr}}, $v_t(X_t|\rvy_0)$ is obtained by solving the modified LQR problem~\eqref{eq:gen-lqr-rev}:
\begin{align}
    \label{eq:gen-lqr-rev}
    V(c) =  \int_0^1 \frac{1}{2}\left\|c\left(Z_t, t\right)\right\|_2^2 \deriv t + \frac{\lambda}{2} \left\|Z_1 - \rvy_0\right\|_2^2, ~\deriv Z_t = c\left(Z_t,t\right) \deriv t, \quad Z_0 =\rvy_1.
\end{align}
Solving \eqref{eq:gen-lqr-rev}, we get $c(Z_t,t) = \frac{\rvy_0 - Z_t}{1-t}$.
Our controller steers the samples toward the given image $\rvy_0$.
Thus, the controlled reverse ODE~\eqref{eq:gen-ode-w-controller} effectively reduces the reconstruction error incurred in the standard reverse ODE~\eqref{eq:gen-ode} of RF models (e.g. Flux).

\textbf{Reverse process using SDE.} Finally, in \textbf{Theorem~\ref{theorem:sde-equiv-cntrl-gen-ode}}, we provide the stochastic equivalent of our controlled reverse ODE~\eqref{eq:gen-ode-w-controller} for generation. Recall that we initialize with the terminal structured noise by running the controlled forward ODE~\eqref{eq:controlled-ODE}, along with a reference image $\rvy_0$. As discussed above, we terminate the inversion process at a time $T = 1 - \delta$ for numerical stability, resulting in a vector $\rvy_{1-\delta}$. Our reverse SDE thus starts at a corresponding time $\delta$ with this vector $\rvy_{1-\delta}$ at initialization, and terminates at time $T' < 1.$  

\begin{theorem}
\label{theorem:sde-equiv-cntrl-gen-ode}
Fix any $T' \in (\delta,1)$, and for any $t \in [\delta, T']$, the density evolution of the controlled ODE~\eqref{eq:gen-ode-w-controller} initialized at $X_0 = \rvy_{1-\delta}$ is identical to the density evolution of the following SDE:
\begin{align}
    \label{eq:stoch-rect-flow-cond-sampling}
    \deriv X_t = \left[\frac{(1-t-\eta)X_t+\eta t \rvy_0}{t(1-t)} + \frac{2(1-t)(1-\eta)}{t}  \nabla \log p_{1-t}(X_t) \right] \deriv t + \sqrt{\frac{2(1-t)(1-\eta)}{t}}  \deriv W_t.
\end{align}
Furthermore, denoting $q_t(\cdot)$ as the marginal pdf of $X_t$, its density evolution is given by: 
\begin{align}
    \frac{\partial q_t(X_t)}{\partial t} = \nabla \cdot \left[-\left(\frac{1-t-\eta}{t(1-t)}X_t +  \frac{\eta}{1-t}\rvy_0 + \frac{(1-t)}{t} (1-\eta) \nabla \log p_{1-t}(X_t)\right)q_t(Y_t)\right].
\end{align}
\end{theorem}

\noindent\textbf{Properties of SDE~\eqref{eq:stoch-rect-flow-cond-sampling}.}
When the controller parameter $\eta = 0$, we obtain a stochastic sampler \eqref{eq:stoch-rect-flow-sampling} for the pre-trained Flux, as given in \textbf{Lemma~\ref{lemma:stoch-rect-flow-sampling}} and compared qualitatively in \Figref{fig:appl-resde}.
This case of our SDE~\eqref{eq:stoch-rect-flow-cond-sampling} corresponds to the stochastic variant of standard RFs~\citep{rectflow,lipman2022flow,interpolant}.
Our key contribution lies in conditioning on $X_1=\rvy_0$ for inverting rectified flows. 
Importantly, our explicit construction does not require additional training or test-time optimization, enabling
for the first time an efficient sampler for zero-shot inversion and editing using Flux.
When $\eta = 1$, the score term and Brownian motion vanish from the SDE~\eqref{eq:stoch-rect-flow-cond-sampling}. The resulting drift becomes $\frac{\rvy_0 - X_t}{1-t}$, the optimal controller for the LQR problem~\eqref{eq:gen-lqr-rev}, exactly recovering the given image $\rvy_0$.

\begin{remark}
\label{rmk:controlled-sde}
Similar to \textbf{Remark~\ref{rmk:controlled-ode}}, our analysis extends to the case when $\eta$ is time-varying.
This is useful in editing, as it allows the flow to initially move toward the given image $\rvy_0$ by choosing a large $\eta$. 
As the flow approaches $\rvy_0$ on the image manifold, $\eta$ is gradually reduced, ensuring that the text-guided edits are enforced through the unconditional vector field $v_t(\cdot)$ provided by Flux.
\end{remark}

\vspace{-2ex}
\section{Algorithm: Inversion and Editing via Controlled ODEs}
\label{sec:algos}
\vspace{-1ex}
We describe the algorithm for RF inversion and editing using our controlled ODEs \eqref{eq:controlled-ODE} and \eqref{eq:gen-ode-w-controller}.

\textbf{Problem Setup.} 
The user provides a text ``prompt" to edit reference content, which could be a corrupt or a clean image. 
For the corrupt image guide, we use the dataset from SDEdit~\citep{sdedit}, which contains color strokes to convey high-level details.
In this setting, the reference guide $\rvy_0$ is typically not a realistic image under the data distribution $p_0$.
The objective is to transform this guide into a more realistic image under $p_0$ while maintaining faithfulness to the original guide. 

For the clean image guide, the user provides a real image $\rvy_0$ along with an accompanying text ``prompt" to specify the desired edits.
The task is to apply text-guided edits to $\rvy_0$ while preserving its content. 
Examples include face editing, where the text might instruct change in age or gender.

\textbf{Procedure.} Our algorithm has two key steps: \textbf{inversion} and \textbf{editing}. We discuss each step below.

\begin{wrapfigure}{r}{0.3\textwidth}
\includegraphics[width=\linewidth]{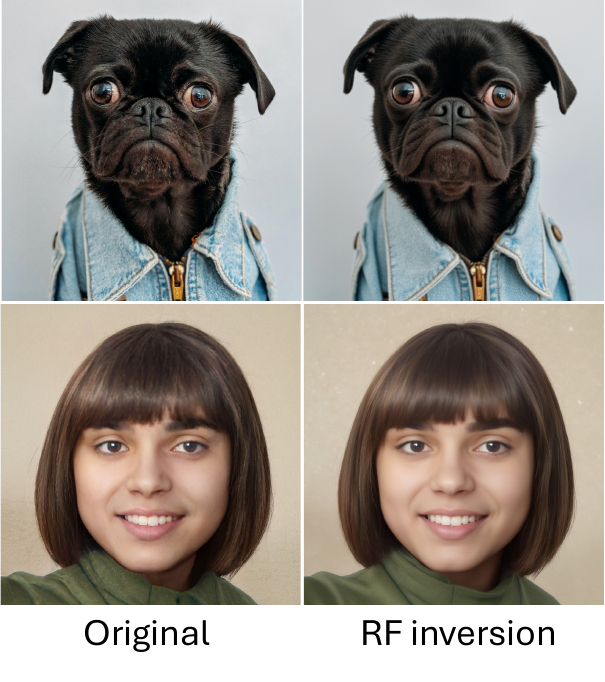}
\caption{
Inverting flows by controlled ODEs~\eqref{eq:controlled-ODE} and \eqref{eq:gen-ode-w-controller}.
}
\vspace{-2ex}  
\label{fig:rf-inv-teaser}
\end{wrapfigure}
\textbf{Inversion.} The first step involves computing the structured noise $\rvy_1$ by employing our controlled ODE~\eqref{eq:controlled-ODE}, initialized at the reference content $Y_0 = \rvy_0$. 
To compute the unconditional vector field, we use the pre-trained Flux model $u\left(\cdot,\cdot,\cdot;\varphi \right)$, which requires three inputs: the state $Y_t$, the time $t$, and the prompt embedding $\Phi(\text{prompt})$.
During the inversion process, we use null prompt in the Flux model, i.e., $u_t(\rvy_t) = u(\rvy_t, t,\Phi(\text{``"});\varphi)$.
For the conditional vector field, we apply the analytical solution derived in \textbf{Proposition~\ref{prop:rf-lqr}}.
The inversion process yields a latent variable that is then used to initialize our controlled ODE~\eqref{eq:gen-ode-w-controller}, i.e., $X_0 = \rvy_1$.
In this phase, we again use the null prompt to compute the vector field $v_t(\rvx_t) = -u(\rvx_t, 1-t,\Phi(\text{``"});\varphi)$: see \Figref{fig:rf-inv-teaser} for the final output.

\textbf{Editing.} The second step involves text-guided editing of the reference content $\rvy_0$. This process is governed by our controlled ODE~\eqref{eq:gen-ode-w-controller}, where the vector field is computed using the desired text prompt within Flux: $v_t(X_t)= -u(\rvx_t, 1-t,\Phi(\text{prompt});\varphi)$.
The controller guidance $\eta$ in \eqref{eq:gen-ode-w-controller} balances faithfulness and editability: higher $\eta$ improves faithfulness but limits editability, while lower $\eta$ allows significant edits at the cost of reduced faithfulness. Consequently, the controller guidance $\eta$ provides a smooth interpolation between faithfulness and editability, a crucial feature in semantic image editing.
Motivated by \textbf{Remark~\ref{rmk:controlled-ode}} and \textbf{\ref{rmk:controlled-sde}}, we consider a time-varying controller guidance $\eta_t$, such that for a fixed $\eta \in [0,1]$ and $\tau \in [0,1]$, $\eta_t = \eta ~\forall t \leq \tau$ and $0$ otherwise.
\Figref{fig:control-strength} illustrates the effect of controller guidance  $\eta$ for $\tau = 0.3$; see Appendix~\ref{sec:ablation} for a detailed ablation study. 
\vspace{-2ex}

\begin{figure}[!t]
    \vspace{-1ex}
    \includegraphics[width=\linewidth]{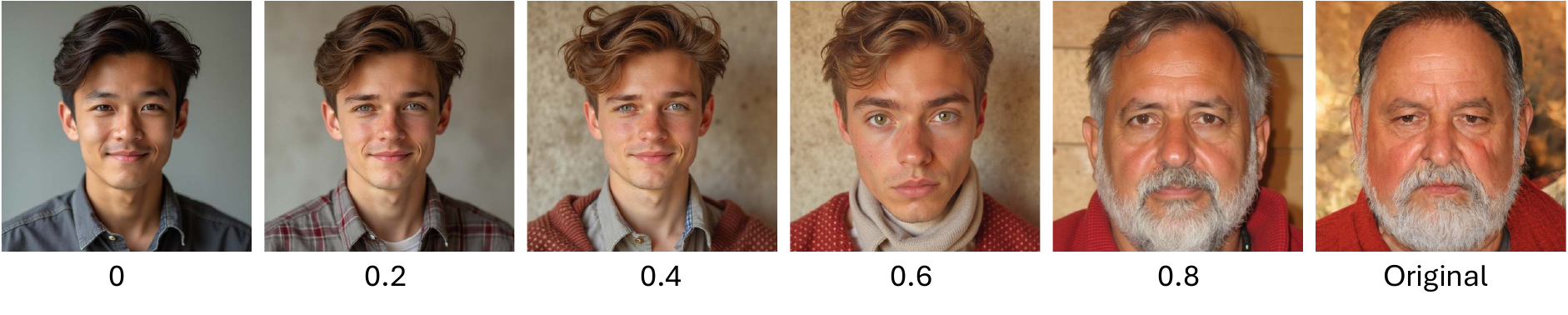}
    \vspace{-3ex}
    \caption{
    \textbf{Effect of controller guidance $\eta$} given the original image and the prompt: ``A young man". 
    Increasing $\eta$ improves the faithfulness to the original image, which is reconstructed at $\eta\!=\!1$.
    }
    \label{fig:control-strength}
    \vspace{-2ex}
\end{figure}

\begin{figure}[!t]
    \vspace{-1ex}
    \includegraphics[width=\linewidth]{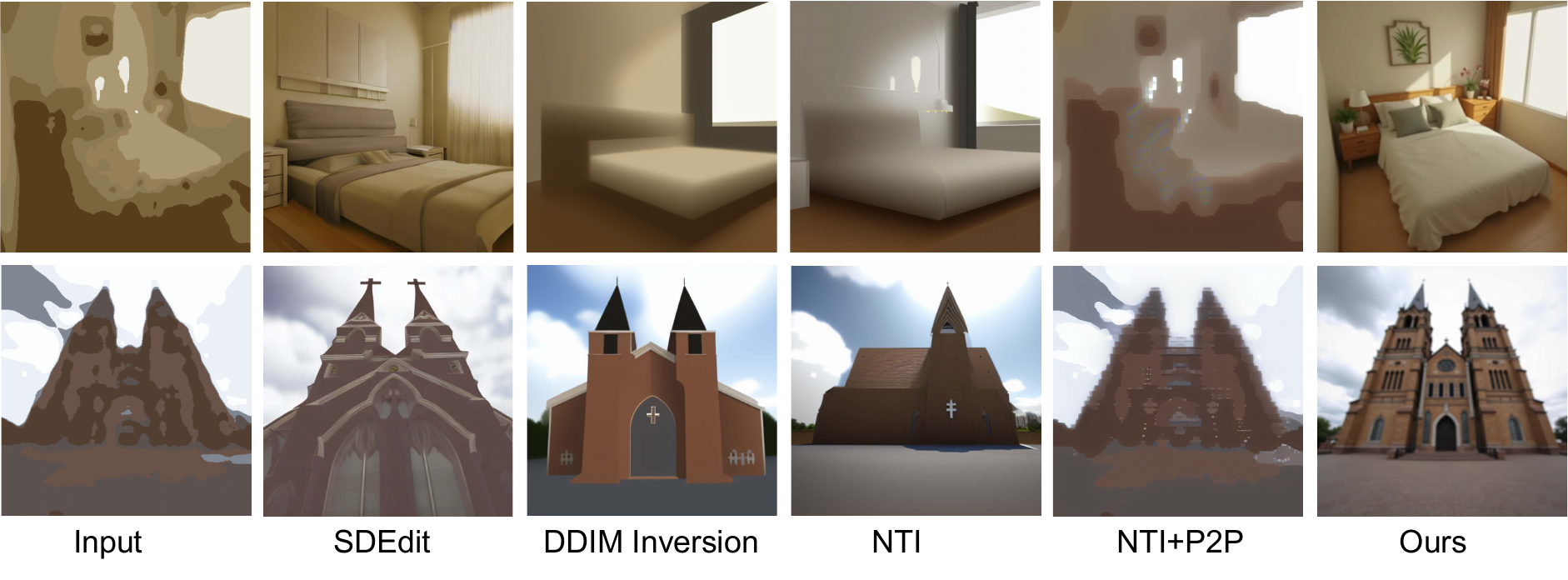}
    \vspace{-3ex}
    \caption{
    \textbf{Stroke2Image generation.} 
    Our method generates photo-realistic images of bedroom or church given stroke paints, showing robustness to initial corruptions.
    }
    \label{fig:stroke2image}
    \vspace{-1ex}  
\end{figure}

\begin{figure}[!t]
    \vspace{-1ex}
    \includegraphics[width=\linewidth]{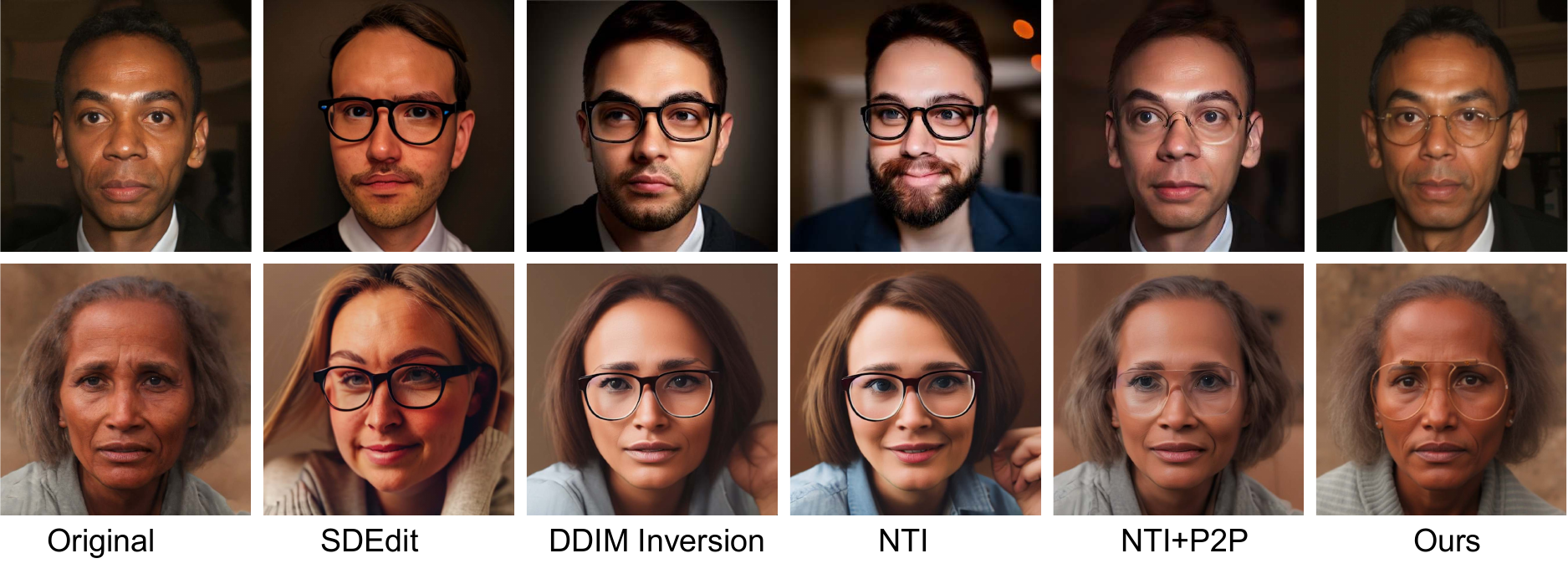}
    \vspace{-4ex}
    \caption{
    \textbf{Image editing for adding face accessories.} 
     Prompt: ``face of a man/woman wearing glasses".
     The proposed method better preserves the identity while applying the desired edits.
     \vspace{-3ex}
    }
    \label{fig:glass-edit}
    \vspace{-1ex}
\end{figure}

\section{Experimental Evaluation}
\label{sec:exps}
\vspace{-1ex}
We show that RF inversion outperforms DM inversion across three benchmarks: LSUN-church, LSUN-bedroom~\citep{lsun}, and SFHQ~\citep{sfhq} on two tasks: Stroke2Image generation and semantic image editing.
Stroke2Image generation shows the robustness of our algorithm to initial corruption.
In semantic image editing, we emphasize the ability to edit clean images without additional training, optimization, or complex attention processors.

\textbf{Baselines.} 
As this paper focuses on inverting flows, we compare with SoTA inversion approaches, such as NTI~\citep{nti}, DDIM Inversion~\citep{ddim}, and SDEdit~\citep{sdedit}. 
We use the official NTI implementation for both NTI and DDIM inversion, and Diffusers library for SDEdit.
Hyper-parameters for all these baselines are tuned for optimal performance.
We compare with NTI for both direct prompt change and with prompt-to-prompt~\cite{p2p} editing.
All methods are training-free; however, NTI~\citep{nti} solves an optimization problem at each denoising step during inversion and uses P2P~\citep{p2p} attention processor during editing.
We follow the evaluation protocol from SDEdit~\citep{sdedit}. More qualitative results and comparison are in Appendix \S\ref{sec:addn-exp}.

\textbf{Stroke2Image generation.}
As discussed in \S\ref{sec:algos}, our goal is to generate a photo-realistic image from a stroke paint (a corrupted image) and the text prompt ``photo-realistic picture of a bedroom".
In this case, the high level details in the stroke painting guide the reverse process toward a clean image. 

In \Figref{fig:stroke2image}, we compare RF inversion (ours) with DM inversions.
DM inversions propagate the corruption from the stroke painting into the structured noise,
which leads to outputs resembling the input stroke painting. 
NTI optimizes null embeddings to align the reverse process with the DDIM forward trajectory. 
Although adding P2P to the NTI pipeline helps localized editing as in Figure~\ref{fig:glass-edit}, for corrupted images, it drives the reverse process even closer to the corruption.
In contrast, our controlled ODE~\eqref{eq:controlled-ODE} yields a structured noise that is consistent with the corrupted image and also the invariant terminal distribution, as discussed in \S\ref{sec:rf-inv}, resulting in more realistic images.

In Table~\ref{tab:lsun-both}, we show that our method outperforms prior works in faithfulness and realism. 
On the test split of LSUN bedroom dataset, our approach is 4.7\% more faithful and 13.79\% more realistic than the best optimization free method SDEdit-SD1.5. Ours is 73\% more realistic than the optimization-based method NTI, but comparable in L2. As discussed, NTI+P2P gets closer toward the corrupt image, which gives a very low L2 error, but the resulting image becomes unrealistic. Our approach is 89\% more realistic than NTI+P2P. We observe similar gains on LSUN church dataset.

\textbf{User study.}
We conduct a user study using Amazon Mechanical Turk to evaluate the overall performance of the our method. 
With 3 responses for each question, we collected in total 9,000 comparisons from 126 participants. 
As given in Table~\ref{tab:lsun-both}, our method outperforms all the other baselines by at least 59.67\% in terms of overall satisfaction.
More details are provided in Appendix \S\ref{sec:human-eval}.

\begin{figure}[!t]
\vspace{-6ex}  
\includegraphics[width=\linewidth]{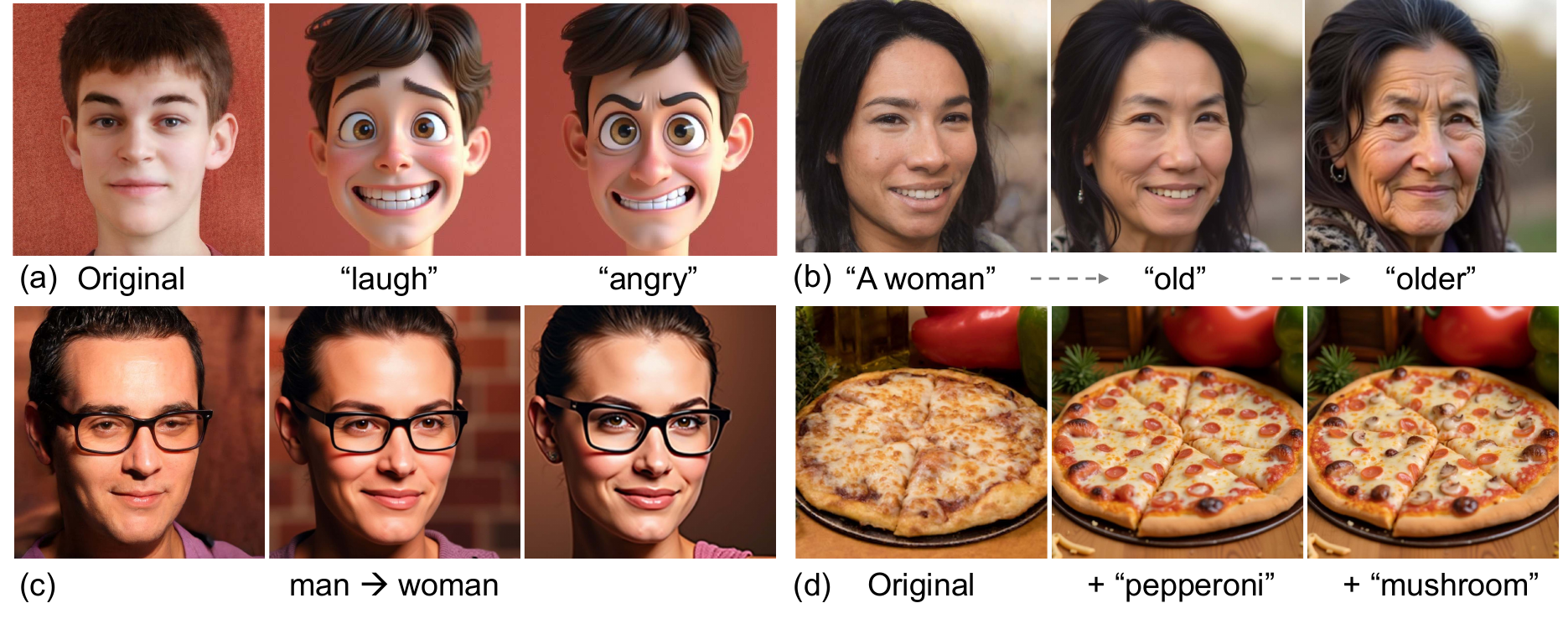}
\caption{
Editing (a) stylized expression, (b) age, (c) gender, and (d) object insert.
Given an original image and a text prompt, our algorithm performs semantic image editing in the wild.
\vspace{-1ex}
}
\label{fig:sem-edit-all}
\vspace{-3ex}
\end{figure}

\begin{table}[!t]
\vspace{-2ex}
\centering
\caption{
\textbf{Quantitative results for Stroke2Image generation.}
L2 and Kernel Inception Distance (KID) capture faithfulness and realism, respectively. Optimization-based methods are colored gray. User Pref. shows the percentage of users that prefer our method over each alternative in pairwise comparisons (and ties). E.g.: 62.11\% (+ 8\% ties) prefer ours over SDEdit-Flux for LSUN Bedroom.
}
\label{tab:lsun-both}
\small
\newlength{\ts}\setlength{\ts}{5mm}
\newcommand{\tts}{\hspace{\ts}}
\begin{tabular}{l*{3}{@{\tts}c}@{\tts}l*{2}{@{\tts}c}}
\toprule
\textbf{} & \multicolumn{3}{c}{\textbf{LSUN Bedroom}} & \multicolumn{3}{c}{\textbf{LSUN Church}} \\
\cmidrule(lr){2-4} \cmidrule(lr){5-7}
{\bf Method} & \textbf{L2} $\downarrow$ & \textbf{KID} $\downarrow$ & \textbf{User Pref. (\%)} $\uparrow$ & \textbf{L2} $\downarrow$ & \textbf{KID} $\downarrow$ & \textbf{User Pref. (\%)} $\uparrow$\\
\midrule
SDEdit-SD1.5  & 86.72 & 0.029 & 59.67 (5.33) & 90.72 & 0.089 & 65.33 (4.11) \\ 
SDEdit-Flux   & 94.89 & 0.032 & 62.11 (8.00) & 92.47 & 0.081 & 66.22 (5.22) \\ 
DDIM Inv. & 87.95 & 0.113 & 82.56 (1.67) & 97.36 & 0.107 & 85.44 (2.78) \\ 
\rowcolor{gray!10}
NTI   & 82.77 & 0.095 & 80.89 (4.33) & 87.88 & 0.098 & 77.11 (4.89) \\ 
\rowcolor{gray!10}
NTI+P2P    &  46.46     &   0.234    & 98.11 (1.78) &   34.48    &  0.168     & 99.22 (0.56) \\ 
\rowcolor{orange!25}
Ours          & {82.65} & {0.025} & - & {80.36} & {0.059} &  -\\ 
\bottomrule
\end{tabular}
\vspace{-2ex}
\end{table}

\textbf{Semantic Image Editing.}
Given a \textit{clean image} and a text ``prompt", the objective is to modify the image according to the given text while preserving the contents of the image (identity for face images). 
In rectified linear paths, editing from a noisy latent becomes straightforward, further enhancing the efficiency of our approach.
Compared with SoTA approaches (\Figref{fig:glass-edit}),
our method requires no additional optimization or complex attention processors as in NTI~\citep{nti}+P2P\citep{p2p}.
Thus, it is more efficient than a current SoTA approach, and importantly, more faithful to the original image while applying the desired edits.

\begin{wrapfigure}{r}{0.5\textwidth}  
\vspace{-2ex}  
\includegraphics[width=\linewidth]{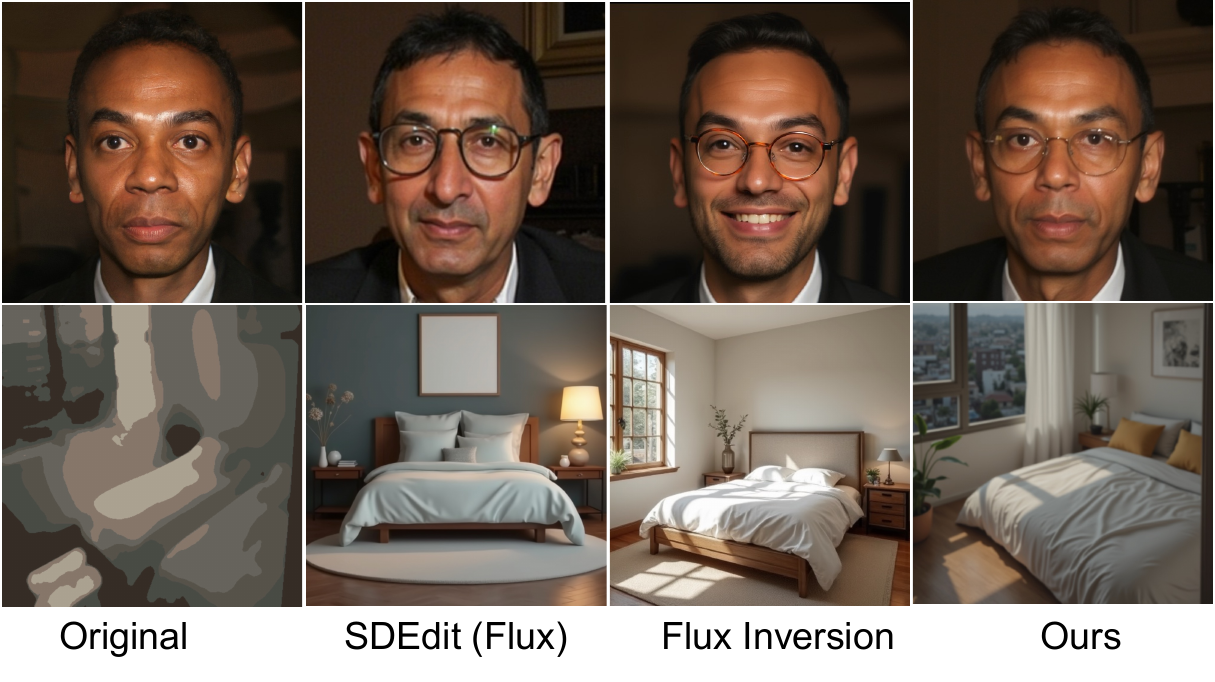}
\caption{
Comparison using Flux backbone.
}
\label{fig:comp-flix-bb}
\vspace{-2ex}  
\end{wrapfigure}
In Table~\ref{tab:sfhq}, we show that our method outperforms the optimization-free methods by at least 29\% in face reconstruction, 6.6\% in DINO patch-wise similarity, and 26.4\% in CLIP-Image similarity while being comparable in prompt alignment metric CLIP-T. 
Importantly, our approach offers 54.11\% gain in runtime, though it uses a larger ($\sim$12X) model, while staying comparable to NTI+P2P.

In \Figref{fig:sem-edit-all}, we showcase four complex editing tasks: 
(a) prompt-based stylization with the prompt: ``face of a boy in disney 3d cartoon style", where facial expressions, such as ``laugh" or ``angry" are used for editing;
(b) ability to control the age of a person;
(c) interpolating between two concepts: ``A man" $\leftrightarrow$ ``A woman";
(d) sequentially inserting pepperoni and mushroom to an image of a pizza.
We provide more examples of editing in the wild in Appendix~\S\ref{sec:addn-exp}.

\textbf{Comparison using the same backbone: Flux.}
In \Figref{fig:comp-flix-bb}, we compare our method with SDEdit and DDIM inversion both adapted to Flux.
NTI optimizes null embeddings to align with forward latents before applying text-guided edits via P2P, an approach well-suited for DMs that use both null and text embedding. 
However, this strategy cannot be applied to Flux, as it does not explicitly use null embedding. 
Consequently, we only reimplement SDEdit and DDIM inversion for Flux and compare them to our method.
Since all methods leverage the same generative model, the improvements clearly stem from our controlled ODEs, grounded in a solid theoretical foundation (\S\ref{sec:method}). 

\begin{table}[!t]
\vspace{-2ex}
\centering
\caption{
\textbf{Quantitative results for face editing} on SFHQ for ``wearing glasses".
}
\label{tab:sfhq}
\small
\begin{tabular}{lccccc}
\toprule
\textbf{Method} & \textbf{Face Rec.} $\downarrow$ & \textbf{DINO} $\uparrow$ & \textbf{CLIP-T} $\uparrow$ & \textbf{CLIP-I} $\uparrow$ & \textbf{Runtime(s)} $\downarrow$\\ 
\midrule
SDEdit-SD1.5 & 0.626 & 0.885 & 0.300  & 0.712 & 8\\ 
SDEdit-Flux & 0.632 & 0.892 & 0.292 & 0.710 & 24\\ 
DDIM Inv. & 0.709 & 0.884 & 0.311 & 0.669 & 15\\ 
\rowcolor{gray!10}
NTI & 0.707 & 0.876 & 0.304 & 0.666 & 78\\ 
\rowcolor{gray!10}
NTI+P2P & 0.443 & 0.953 & 0.293 & 0.845 & 85\\ 
\rowcolor{orange!25}
Ours & {0.442} & {0.951}  & 0.300  & 0.900 & 39\\ 
\bottomrule
\end{tabular}
\vspace{-3ex}
\end{table}

\section{Conclusion}
\label{sec:conc}
\vspace{-1ex}
We present the first \textit{efficient} approach for inversion and editing with the state-of-art rectified flow models such as Flux. Our method interpolates between two vector fields: {\em (i)} the unconditional RF field that transforms a ``clean'' image to  ``typical'' noise, and {\em (ii)} a conditional vector field derived from optimal control that transforms {\em any} image (clean or not) to ``typical'' noise. Our new field thus navigates between these two competing objectives of {\em consistency with the given (possibly corrupted) image, and consistency with the distribution of clean images}. Theoretically, we show that this is equivalent to a new rectified SDE formulation, sharing this intuition of interpolation. Practically, we show that our method results in state-of-art zero-shot performance, without the need of additional training, optimization of latent variables,  prompt tuning, or complex attention processors.
We demonstrate the effectiveness of our method in stroke-to-image synthesis, face editing, object insertion, and stylization tasks, with large-scale human evaluation confirming user preference.

\noindent\textbf{Limitation.} 
The lack of comparison with \textit{expensive} diffusion-based editing solutions may be viewed as a limitation. 
However, these implementations are either not available for Flux or not directly applicable due to Flux's distinct multi-modal architecture. 
The key contribution of this paper lies in its theoretical foundations, validated using standard benchmarks and relevant baselines.

\noindent\textbf{Reproducibility.} The pseudocode and hyper-parameter details have been provided to reproduce the reported results in this paper. 

\section*{Broader Impact Statement}
\label{sec:broad-impact}
Semantic image inversion and editing have both positive and negative social impacts.

On the positive side, this technology enables (i) the generation of photo-realistic images from high level descriptions, such as stroke paintings, and (ii) the modification of clean images by changing various attributes like the age, gender, or adding glasses (\S\ref{sec:exps}). 

On the negative side, it can be misused by malicious users to manipulate photographs of individuals with inappropriate or offensive edits. 
Additionally, it carries the inherent risks associated with the underlying generative model.

To mitigate the negative social impacts, we enable safety features such as NSFW filters in the underlying generative model. 
Furthermore, we believe watermarking images generated by this technology can reduce misuse, especially in inversion and editing applications. 

\section*{Acknowledgments}
This research has been supported by NSF Grant 2019844, a Google research collaboration award, and the UT Austin Machine Learning Lab.

\bibliography{iclr2025_conference}
\bibliographystyle{iclr2025_conference}

\clearpage
\pagebreak
\appendix
\section{Additional Theoretical Results}
\label{sec:addn-theory}
In this section, we present the theoretical results omitted from the main draft due to space constraints.
We formalize the ODE derivation of the standard rectified flows in \textbf{Lemma~\ref{lemma:optimal-vector-field}}.
\begin{lemma}
\label{lemma:optimal-vector-field}
Given a coupling $(Y_0, Y_1) \sim p_0 \times p_1$, consider the noising process $Y_t = tY_1 + (1-t)Y_0$.
Then, the rectified flow ODE formulation with the optimal vector field is given by 
\begin{align}
\label{eq:ode-optimal-vector-field}
    \deriv Y_t = \left[ -\frac{1}{1-t}Y_t - \frac{t}{1-t} \nabla \log p_t(Y_t) \right]\deriv t,\quad Y_0 \sim p_0.
\end{align}
Furthermore, denoting $p_t(\cdot)$ as the marginal pdf of $Y_t$, its density evolution is given by: 
\begin{align}
    \label{eq:ode-optimal-vf-fpe}
    \frac{\partial p_t(Y_t)}{\partial t}= \nabla \cdot \left[\left( \frac{1}{1-t}Y_t + \frac{t}{1-t}~\nabla \log p_t(Y_t) \right)p_t(Y_t)\right].
\end{align}
\end{lemma}
In \textbf{Lemma~\ref{lemma:sde2ode}}, we show that the marginal distribution of the rectified flow~\eqref{eq:inv-wo-control} is equal to that of an SDE with appropriate drift and diffusion terms.
\begin{lemma}
\label{lemma:sde2ode}
Fix any $T \in (0, 1)$, and for any $t \in [0, T]$, the density evolution \eqref{eq:ode-optimal-vf-fpe} of the rectified flow model \eqref{eq:ode-optimal-vector-field} is identical to the density evolution of the following SDE:
\begin{align}
    \label{eq:app-sde-optimal-vector-field}
    \deriv Y_t = -\frac{1}{1-t}Y_t \deriv t + \sqrt{\frac{2t}{1-t}} \deriv W_t, \quad Y_0\sim p_0.
\end{align}
\end{lemma}
In \textbf{Proposition~\ref{prop:rect-flow-ode-equals-sde}}, we show that the stationary distribution of the SDE~\eqref{eq:app-sde-optimal-vector-field} converges to the standard Gaussian $\gN(0,I)$ in the limit as $t\rightarrow 1$.
\begin{proposition}
\label{prop:rect-flow-ode-equals-sde}
Fix any $T \in (0, 1)$, and for any $t \in [0, T]$, the density evolution for the rectified flow ODE \eqref{eq:inv-wo-control} is same as that of the SDE~\eqref{eq:sde-optimal-vector-field}.
Furthermore, denoting $p_t(\cdot)$ as the marginal pdf of $Y_t$, its stationary distribution $p_t(Y_t) \propto \exp{(-\frac{\left\|Y_t\right\|^2}{2t})}$, which converges to $\gN\left(0,I\right)$ as $t\rightarrow1$.
\end{proposition}

We note that Lemma~\ref{lemma:optimal-vector-field} and Lemma~\ref{lemma:sde2ode} follow from the duality between the heat equation and the continuity equation~\citep{oksendal2003stochastic}, where it is classically known that one can interpret a diffusive term as a vector field that is affine in the score function, and vice-versa. This connection has been carefully used to study a large family of stochastic interpolants (that generalize rectified flows) in \citep{interpolant, interpolants2}, and which can lead to a family of ODE-SDE pairs. In the lemmas above, we have provided explicit coefficients that have been directly derived, instead of using the stochastic interpolant formulation.
Our key contribution lies in constructing a controlled ODEs~\eqref{eq:controlled-ODE} and \eqref{eq:gen-ode-w-controller}, along with their equivalent SDEs~\eqref{eq:sde-controlled-vector-field} and \eqref{eq:stoch-rect-flow-cond-sampling} in Theorem~\ref{theorem:sde-equiv-cntrl-ode} and Theorem~\ref{theorem:sde-equiv-cntrl-gen-ode}, respectively.
This aids faithfulness and editability as discussed in \S\ref{sec:algos}.

In \textbf{Lemma~\ref{lemma:stoch-rect-flow-sampling}}, we derive a rectified SDE that transforms noise into images by reversing the stochastic equivalent of rectified flows~\eqref{eq:sde-optimal-vector-field}.

\begin{lemma}
\label{lemma:stoch-rect-flow-sampling}
Fix any small $\delta \in (0, 1)$, and for any $t \in [\delta, 1]$, the process $X_t$ governed by the SDE:
\begin{align}
    \label{eq:stoch-rect-flow-sampling}
    \deriv X_t = \left[\frac{1}{t}X_t + \frac{2(1-t)}{t} \nabla \log p_{1-t}(X_t) \right] \deriv t + \sqrt{\frac{2(1-t)}{t}} \deriv W_t, \quad X_0 \sim p_1,
\end{align}
is the time-reversal of the SDE~\eqref{eq:sde-optimal-vector-field}.
\end{lemma}
\textbf{Implication.}
The reverse SDE~\eqref{eq:stoch-rect-flow-sampling} provides a stochastic sampler for SoTA rectified flow models like Flux.
Unlike diffusion-based generative models that explicitly model the score function $\nabla \log p_{t}(\cdot)$ in \eqref{eq:stoch-rect-flow-sampling}, rectified flows model a vector field, as discussed in \S\ref{sec:prelim}.
However, given a neural network $u(\rvy_t,t;\varphi))$ approximating the vector field $u_t(\rvy_t)$, \textbf{Lemma~\ref{lemma:optimal-vector-field}} offers an explicit formula for computing the score function:
\begin{align}
    \label{eq:flux-to-score}
    \nabla \log p_t(Y_t) = -\frac{1}{t} Y_t - \frac{1-t}{t}u(Y_t,t;\varphi).
\end{align}
This score function is used to compute the drift and diffusion coefficients of the SDE~\eqref{eq:stoch-rect-flow-sampling},
resulting in a practically implementable stochastic sampler for Flux.
This extends the applicability of Flux to downstream tasks where SDE-based samplers have demonstrated practical benefits, as seen in diffusion models~\citep{ddpm,songscore,ldm,sdxl}.

\section{Technical Proofs}
\label{sec:proofs}
This section contains technical proofs of the theoretical results presented in this paper.
\subsection{Proof of Proposition~\ref{prop:rf-lqr}}
\label{sec:proofs-prop-rf-lqr}
\begin{proof}
The standard approach to solving an LQR problem is the minimum principle theorem that can be found in control literature~\citep{fleming2012deterministic,basar2020lecture}. We follow this approach and provide the full proof below for completeness.

The Hamiltonian of the LQR problem~\eqref{eq:gen-lqr} is given by
\begin{align}
    H(\rvz_t, \rvp_t, \rvc_t, t) = \frac{1}{2}\left\|\rvc_t\right\|^2 + \rvp_t^T\rvc_t.
\end{align}
For $\rvc_t^* = -\rvp_t$, the Hamiltonian attains its minumum value: $H(\rvz_t, \rvp_t, \rvc_t^*, t) = -\frac{1}{2}\left\|\rvp_t \right\|^2$.
Using minimum principle theorem~\citep{fleming2012deterministic,basar2020lecture}, we get
\begin{align}
\label{eq:lqr-1}
\frac{\deriv \rvp_t}{\deriv t} 
    &= \nabla_{\rvz_t} H\left(\rvz_t, \rvp_t, \rvc_t^*, t\right) = 0;\\
    \label{eq:lqr-2}
    \frac{\deriv \rvz_t}{\deriv t} 
    &= \nabla_{\rvp_t} H\left(\rvz_t, \rvp_t, \rvc_t^*, t\right) = -\rvp_t;\\
    \label{eq:lqr-3}
    \rvz_0 &= \rvy_0;\\
    \label{eq:lqr-4}
    \rvp_1 &= \nabla_{\rvz_1}  \left( \frac{\lambda}{2} \left\|\rvz_1 - \rvy_1\right\|_2^2\right) = \lambda \left(\rvz_1 - \rvy_1 \right).
\end{align}
From \eqref{eq:lqr-1}, we know $\rvp_t$ is a constant $\rvp$. 
Using this constant in \eqref{eq:lqr-2} and integrating from $t\rightarrow1$, we have $\rvz_1 = \rvz_t - \rvp (1-t)$.
Substituting $\rvz_1$ in \eqref{eq:lqr-3}, 
\begin{align*}
    \rvp = \lambda (\rvz_t - \rvp(1-t)-\rvy_1) = \lambda (\rvz_t - \rvy_1) - \lambda (1-t)\rvp,
\end{align*}
which simplifies to 
\begin{align*}
    \rvp = \left(1+\lambda (1-t)\right)^{-1} \lambda (\rvz_t - \rvy_1)
    = \left(\frac{1}{\lambda}+(1-t)\right)^{-1} (\rvz_t - \rvy_1).
\end{align*}
Taking the limit $\lambda \rightarrow \infty$, we get $\rvp = \frac{\rvz_t - \rvy_1}{1-t}$ and the optimal controller $\rvc_t^* = \frac{\rvy_1 - \rvz_t}{1-t}$. Since $u_t(\rvz_t | \rvy_1) = \rvy_1 - \rvy_0$, the proof follows by substituting $\rvy_0 = \frac{\rvz_t - t\rvy_1}{1-t}$. 
\end{proof}

\subsection{Proof of Proposition~\ref{prop:inv-wo-control}}
\label{sec:proofs-prop-inv-wo-control}
\begin{proof}
Initializing the generative ODE~\eqref{eq:gen-ode} with the structured noise $\rvy_1$, we get
\begin{align}
\label{eq:inv-wo-contrl-1}
    \frac{\deriv X_t}{\deriv t} = v_t(X_t), \quad X_0 =\rvy_1, \quad \forall t\in[0,1].
\end{align}
Substituting $u_t(\cdot) = -v_{1-t}(\cdot)$ in ODE~\eqref{eq:inv-wo-control}, 
\begin{align*}
    \frac{\deriv Y_t}{\deriv t} 
    = u_t(Y_t)
    = -v_{1-t}(Y_t),\quad Y_0 = \rvy_0, \quad \forall t\in[0,1].
\end{align*}
Replacing $t\rightarrow (1-t)$,
\begin{align}
\label{eq:inv-wo-contrl-2}
    \frac{\deriv Y_{1-t}}{\deriv t} 
    = v_{t}(Y_{1-t}), \quad \forall t\in[0,1].
\end{align}
Since \eqref{eq:inv-wo-contrl-1} and \eqref{eq:inv-wo-contrl-2} hold $\forall t\in[0,1]$ and $X_0 = \rvy_1$, then $X_t = Y_{1-t}$ that implies $X_1 = Y_0 = \rvy_0$.
\end{proof}

\subsection{Proof of Theorem~\ref{theorem:sde-equiv-cntrl-ode}}
\label{sec:proofs-thm-sde-equiv-cntrl-ode}
\begin{proof}
From \textbf{Proposition~\ref{prop:rf-lqr}}, we have $u_t(Y_t|Y_1) = \frac{Y_1 - Y_t}{1-t}$. In \textbf{Lemma~\ref{lemma:optimal-vector-field}}, we show that
\begin{align*}
    u_t(Y_t) = \left[ -\frac{1}{1-t}Y_t - \frac{t}{1-t} \nabla \log p_t(Y_t) \right].
\end{align*}
Now, the controlled ODE~\eqref{eq:controlled-ODE} becomes:
\begin{align*}
    \deriv Y_t 
    & = \Big[u_t(Y_t) + \gamma \left(u_t(Y_t|Y_1) - u_t(Y_t)\right) \Big] \deriv t, \quad Y_0 \sim p_0, \quad Y_1 = \rvy_1\\
    & =  \left[(1-\gamma)\left( -\frac{1}{1-t}Y_t - \frac{t}{1-t} \nabla \log p_t(Y_t)\right) + \gamma \left(\frac{Y_1 - Y_t}{1-t}\right)\right]\deriv t\\
    & =  \left[-\frac{1}{1-t}Y_t - \frac{t}{1-t} (1-\gamma)\nabla \log p_t(Y_t) + \frac{\gamma}{1-t}Y_1\right]\deriv t\\
    & =  \left[-\frac{1}{1-t}\left(Y_t - \gamma Y_1\right) - \frac{t}{1-t}(1-\gamma) \nabla \log p_t(Y_t)\right]\deriv t.
\end{align*}
Using continuity equation~\citep{oksendal2003stochastic}, the density evolution of the controlled ODE~\eqref{eq:controlled-ODE} then becomes:
\begin{align}
\label{eq:sde-controlled-vector-field-1}
    \frac{\partial p_t(Y_t)}{\partial t} = \nabla \cdot \left[\left( \frac{1}{1-t}\left(Y_t - \gamma Y_1\right) + \frac{t}{1-t} (1-\gamma)\nabla \log p_t(Y_t)\right)p_t(Y_t)\right].
\end{align}
Applying Fokker-Planck equation~\citep{oksendal2003stochastic} to the SDE~\eqref{eq:sde-controlled-vector-field}, we have
\begin{align*}
    \frac{\partial p_t(Y_t)}{\partial t} 
    + \nabla \cdot \left[\left(-\frac{1}{1-t}\left( Y_t -\gamma Y_1\right) \right) p_t(Y_t) \right] 
    = \nabla \cdot \left[ \frac{t}{1-t}(1-\gamma) \nabla p_t(Y_t)\right],
\end{align*}
which can be rearranged to equal \eqref{eq:sde-controlled-vector-field-1} completing the proof. 
\end{proof}

\subsection{Proof of Lemma~\ref{lemma:optimal-vector-field}}
\label{sec:proofs-lemma-optimal-vector-field}
\begin{proof}
Given $(Y_0, Y_1) \sim p_0 \times p_1$, the conditional flow matching loss~\eqref{eq:cfm-loss} can be reparameterized as:
\begin{align*}
    \gL_{CFM}(\varphi) \coloneqq \E_{t\sim \gU[0,1],(Y_0,Y_1)\sim p_1 \times p_0}\left[\left\| (Y_1 - Y_0) - u(Y_t,t;\varphi) \right\|_2^2\right],\quad Y_t = t Y_1+(1-t)Y_0,
\end{align*}
where the optimal solution is given by the minimum mean squared estimator:
\begin{align}
    \label{eq:ode-optimal-vector-field-1}
    u_t(\rvy_t) = \E_{(Y_0,Y_1)\sim p_1 \times p_0}\left[Y_1 - Y_0 | Y_t=\rvy_t\right].
\end{align}
Since $Y_t = tY_1 + (1-t)Y_0$, we use Tweedie's formula~\citep{efron2011tweedie} to compute
\begin{align}
    \label{eq:ode-optimal-vector-field-2}
    \E\left[Y_0 | Y_t=\rvy_t\right] = \frac{1}{1-t}\rvy_t + \frac{t^2}{1-t} \nabla \log p_t(\rvy_t).
\end{align}
Using the above relation, we obtain the following:
\begin{align}
    \nonumber
    \E\left[Y_1 | Y_t=\rvy_t\right] 
    & = \frac{1}{t} \E\left[Y_t - (1-t)Y_0 | Y_t = \rvy_t\right]\\
    \nonumber
    & = \frac{1}{t} \left(\rvy_t - (1-t) \E\left[Y_0 | Y_t=\rvy_t\right] \right)\\
    \nonumber
    & = \frac{1}{t}\left(\rvy_t - (1-t) \left(\frac{1}{1-t}\rvy_t + \frac{t^2}{1-t} \nabla \log p_t(\rvy_t) \right) \right)\\
    \label{eq:ode-optimal-vector-field-3}
    & = -t ~\nabla \log p_t(\rvy_t).
\end{align}
Combining \eqref{eq:ode-optimal-vector-field-2} and \eqref{eq:ode-optimal-vector-field-3} using linearity of expectation, we get
\begin{align}
    u_t(\rvy_t) 
    & = \E\left[Y_1 | Y_t=\rvy_t\right] - \E\left[Y_0 | Y_t=\rvy_t\right] \\
    & =  -t ~\nabla \log p_t(\rvy_t) - \frac{1}{1-t}\rvy_t - \frac{t^2}{1-t} \nabla \log p_t(\rvy_t)\\
    & = -\frac{1}{1-t} \rvy_t - \frac{t}{1-t} \nabla \log p_t(\rvy_t),
\end{align}
The density evolution of $Y_t$ now immediately follows from the continuity equation \citep{oksendal2003stochastic} applied to \eqref{eq:ode-optimal-vector-field}.
\end{proof}

\subsection{Proof of Lemma~\ref{lemma:sde2ode}}
\label{sec:proofs-lemma-sde2ode}
\begin{proof}
The Fokker-Planck equation of the SDE~\eqref{eq:sde-optimal-vector-field} is given by
\begin{align}
    \label{eq:fokker-planck-controlled-sde}
    \frac{\partial p_t(Y_t)}{\partial t} + \nabla \cdot \left[ - \frac{1}{1-t}Y_t~p_t(Y_t) \right] = \nabla \cdot \left[ \frac{t}{1-t}~\nabla p_t(Y_t) \right].
\end{align}
Rearranging \eqref{eq:fokker-planck-controlled-sde} by multiplying and dividing $p_t(Y_t)$ in the right hand side, we get
\begin{align}
    \label{eq:fokker-planck-controlled-sde-1}
    \frac{\partial p_t(Y_t)}{\partial t} = \nabla \cdot \left[\left( \frac{1}{1-t}Y_t + \frac{t}{1-t}~\nabla \log p_t(Y_t) \right)p_t(Y_t)\right].
\end{align}
To conclude, observe  that that the density evolution above is identical to \eqref{eq:ode-optimal-vf-fpe}. 
\end{proof}

\subsection{Proof of Proposition~\ref{prop:rect-flow-ode-equals-sde}}
\label{sec:proofs-prop-rect-flow-ode-equals-sde}
\begin{proof}
The optimal vector field of the rectified flow ODE~\eqref{eq:inv-wo-control} is given by \textbf{Lemma~\ref{lemma:optimal-vector-field}}. The proof then immediately follows from the Fokker-Planck equations in \textbf{Lemma~\ref{lemma:optimal-vector-field}} and \textbf{Lemma~\ref{lemma:sde2ode}}.

From \textbf{Lemma~\ref{lemma:sde2ode}}, the density evolution of the SDE \eqref{eq:sde-optimal-vector-field} is given by 
 \begin{align*}
    \frac{\partial p_t(Y_t)}{\partial t} = \nabla \cdot \left[\left( \frac{1}{1-t}Y_t + \frac{t}{1-t}~\nabla \log p_t(Y_t) \right)p_t(Y_t)\right].
 \end{align*}
The stationary (or steady state) distribution  satisfies the following:
 \begin{align*}
     \frac{\partial p_t(Y_t)}{\partial t}=0=\nabla \cdot \left[\left( \frac{1}{1-t}Y_t + \frac{t}{1-t}~\nabla \log p_t(Y_t) \right)p_t(Y_t)\right].
 \end{align*}
Using the boundary conditions~\citep{oksendal2003stochastic}, we get
\begin{align*}
    \frac{1}{1-t}Y_t + \frac{t}{1-t}~\nabla \log p_t(Y_t) = 0,
\end{align*}
which immediately implies $p_t(Y_t) \propto e^{-\frac{\left\|Y_t\right\|^2}{2t}}$.
\end{proof}

\subsection{Proof of Theorem~\ref{theorem:sde-equiv-cntrl-gen-ode}}
\label{sec:thm}
\begin{proof}
Using Fokker-Planck equation~\citep{oksendal2003stochastic}, \textbf{Lemma~\ref{lemma:stoch-rect-flow-sampling}} implies
\begin{align*}
    \frac{\partial q_t(X_t)}{\partial t} 
    = \nabla \cdot \left[-q_t(X_t) \left(\frac{1}{t}X_t + \frac{1-t}{t} \nabla \log q_{t}(X_t) \right)\right].
\end{align*}
Therefore, the optimal vector field $v_t(X_t)$ of the controlled ODE~\eqref{eq:gen-ode-w-controller} is given by
\begin{align}
\label{eq:stoch-rect-flow-cond-sampling-1}
    v_t(X_t) = \frac{1}{t}X_t + \frac{1-t}{t} \nabla \log p_{1-t}(X_t).
\end{align}
The LQR problem~\eqref{eq:gen-lqr-rev} is identical to the LQR problem~\eqref{eq:gen-lqr} with changes in the initial and terminal states.
Similar to \textbf{Proposition~\ref{prop:rf-lqr}}, we compute the closed-form solution for the conditional vector field of the ODE~\eqref{eq:gen-ode-w-controller} as:
\begin{align}
\label{eq:stoch-rect-flow-cond-sampling-2}
    v_t(X_t|X_1) = \frac{X_1 - X_t}{1-t}.
\end{align}
Combining \eqref{eq:stoch-rect-flow-cond-sampling-1} and \eqref{eq:stoch-rect-flow-cond-sampling-2}, we have 
\begin{align*}
    \deriv X_t 
    & = \left[v_t(X_t) + \eta (v_t(X_t|X_1) - v_t(X_t))\right]\deriv t\\
    & = \left[(1-\eta)\left(\frac{1}{t}X_t + \frac{1-t}{t} \nabla \log p_{1-t}(X_t)\right) + \eta \left(\frac{X_1 - X_t}{1-t}\right) \right]\deriv t\\
    & = \left[\frac{(1-\eta)(1-t) - \eta t }{t(1-t)}X_t +\frac{\eta}{1-t}X_1 + \frac{(1-\eta)(1-t)}{t}\nabla \log p_{1-t}(X_t)\right] \deriv t\\
    & = \left[\frac{1-t-\eta}{t(1-t)}X_t +\frac{\eta}{1-t}X_1 + \frac{(1-\eta)(1-t)}{t}\nabla \log p_{1-t}(X_t)\right] \deriv t.
\end{align*}
The resulting continuity equation~\citep{oksendal2003stochastic} becomes:
\begin{align*}
    \frac{\partial q_t(X_t)}{\partial t} 
    & = \nabla \cdot \left[-\left( \frac{1-t-\eta}{t(1-t)}X_t +\frac{\eta}{1-t}X_1 + \frac{(1-\eta)(1-t)}{t}\nabla \log p_{1-t}(X_t)\right)q_t(X_t) \right]\\
    & = \nabla \cdot \Bigg[-\left( \frac{1-t-\eta}{t(1-t)}X_t +\frac{\eta}{1-t}X_1 + \frac{2(1-\eta)(1-t)}{t}\nabla \log p_{1-t}(X_t)\right)q_t(X_t)\\
    &\hspace{5cm}+ \left(\frac{(1-\eta)(1-t)}{t}\nabla \log p_{1-t}(X_t)\right)q_t(X_t) \Bigg].
\end{align*}
Using time-reversal property from \textbf{Propsition~\ref{prop:rf-lqr}}, the above expression simplifies to 
\begin{align*}
    \frac{\partial q_t(X_t)}{\partial t} 
    & + \nabla \cdot \Bigg[\left( \frac{1-t-\eta}{t(1-t)}X_t +\frac{\eta}{1-t}X_1 + \frac{2(1-\eta)(1-t)}{t}\nabla \log p_{1-t}(X_t)\right)q_t(X_t) \Bigg] \\
    &= \nabla \cdot \Bigg[\frac{(1-\eta)(1-t)}{t}\nabla q_t(X_t) \Bigg], 
\end{align*}
which yields the following SDE:
\begin{align*}
    \deriv X_t = \left[ \frac{1-t-\eta}{t(1-t)}X_t +\frac{\eta}{1-t}X_1 + \frac{2(1-\eta)(1-t)}{t}\nabla \log p_{1-t}(X_t)\right] \deriv t + \sqrt{\frac{2(1-\eta)(1-t)}{t}} \deriv W_t,
\end{align*}
and thus, completes the proof.
\end{proof}

\subsection{Proof of Lemma~\ref{lemma:stoch-rect-flow-sampling}}
\label{sec:proofs-lemma-stoch-rect-flow-sampling}
\begin{proof}
It suffices to show that the Fokker-Planck equations of the SDE~\eqref{eq:stoch-rect-flow-sampling} and \eqref{eq:sde-optimal-vector-field} are the same after time-reversal. 
Let $q_t(\cdot)$ denote the marginal pdf of $X_t$ such that $q_0(\cdot) = p_1(\cdot)$.
The Fokker-Planck equations of the SDE~\eqref{eq:stoch-rect-flow-sampling} becomes
\begin{align*}
    \frac{\partial q_t(X_t)}{\partial t} + \nabla \cdot \left[q_t(X_t) \left(\frac{1}{t}X_t + \frac{2(1-t)}{t} \nabla \log p_{1-t}(X_t) \right) \right]
    = \nabla \cdot \left[\left(\frac{1-t}{t}\right) \nabla q_t(X_t)\right],
\end{align*}
which can be rearranged to give
\begin{align*}
    \frac{\partial q_t(X_t)}{\partial t} 
    &= \nabla \cdot \left[-q_t(X_t) \left(\frac{1}{t}X_t + \frac{2(1-t)}{t} \nabla \log p_{1-t}(X_t) \right)  
    +
    \left(\frac{1-t}{t}\right) \nabla q_t(X_t)\right]
    \\
    &= \nabla \cdot \left[-q_t(X_t) \left(\frac{1}{t}X_t + \frac{2(1-t)}{t} \nabla \log p_{1-t}(X_t)  
    -
    \frac{1-t}{t} \nabla \log q_t(X_t) \right)\right]
\end{align*}
Since $Y_t$ is the time-reversal process of $X_t$ as discussed in \textbf{Proposition~\eqref{prop:inv-wo-control}},
\begin{align*}
    \frac{\partial q_t(X_t)}{\partial t} 
    = \nabla \cdot \left[-q_t(X_t) \left(\frac{1}{t}X_t + \frac{1-t}{t} \nabla \log q_{t}(X_t) \right)\right].
\end{align*}
Substituting $t\rightarrow 1-t$, 
\begin{align*}
    \frac{\partial q_{1-t}(X_{1-t})}{\partial t} 
    = \nabla \cdot \left[q_{1-t}(X_{1-t}) \left(\frac{1}{1-t}X_{1-t} + \frac{t}{1-t} \nabla \log q_{1-t}(X_{1-t}) \right)\right],
\end{align*}
which implies the density evolution of \eqref{eq:sde-optimal-vector-field}:
\begin{align*}
    \frac{\partial p_{t}(Y_{t})}{\partial t} 
    & = \nabla \cdot \left[p_{t}(Y_{t}) \left(\frac{1}{1-t}Y_{t} + \frac{t}{1-t} \nabla \log p_{t}(Y_{t}) \right)\right].
\end{align*}
This completes the proof of the statement.
\end{proof}

\section{Additional Experiments}
\label{sec:addn-exp}
This section substantiates our contributions further by providing additional experimental details.

\textbf{Baselines.} 
We use the official NTI codebase\footnote{\url{https://github.com/google/prompt-to-prompt}} for the implementations of NTI~\citep{nti}, P2P~\citep{p2p}, and DDIM~\citep{ddim} inversion.
We use the official Diffusers implementation\footnote{\url{https://github.com/huggingface/diffusers}} for SDEdit and Flux\footnote{\url{https://github.com/black-forest-labs/flux}}.
We modify the pipelines for SDEdit and DDIM inversion to adapt to the Flux backbone.

For completeness, we include qualitative comparison with a leading training-based approach InstructPix2Pix~\citep{instructpix2pix}\footnote{\url{https://huggingface.co/spaces/timbrooks/instruct-pix2pix}} and a higher-order differential equation based LEDIT++~\citep{leditspp}\footnote{\url{https://huggingface.co/spaces/editing-images/leditsplusplus}} (\S\ref{sec:addn-exp}).
Table~\ref{tab:baselines} summarizes the requirements of the compared baselines.

\begin{table}[!tbh]
\centering
\caption{
Requirements of compared baselines. 
Our method outperforms prior works while requiring no additional training, optimization of prompt embedding, or attention manipulation scheme. 
}
\label{tab:baselines}
\begin{tabular}{lccc}
\toprule
\textbf{Method} & \textbf{Training} & \textbf{Optimization} & \textbf{Attention Manipulation} \\
\midrule
\rowcolor{orange!25}
SDEdit~\citep{sdedit}  & \tikzxmark & \tikzxmark & \tikzxmark \\
\rowcolor{orange!25}
DDIM~\citep{ddim} & \tikzxmark & \tikzxmark & \tikzxmark \\
\rowcolor{gray!10}
NTI~\citep{nti}  & \tikzxmark & \tikzcmark & \tikzxmark \\
\rowcolor{gray!10}
NTI+P2P~\citep{p2p} & \tikzxmark & \tikzcmark & \tikzcmark \\
LEDIT++~\citep{leditspp} & \tikzxmark & \tikzxmark & \tikzcmark \\
InstructPix2Pix~\citep{instructpix2pix} & \tikzcmark & \tikzxmark & \tikzxmark \\
\rowcolor{orange!25}
Ours & \tikzxmark & \tikzxmark & \tikzxmark \\
\bottomrule
\end{tabular}
\end{table}

\textbf{Metrics.} 
Following SDEdit~\citep{sdedit}, we measure faithfulness using L2 loss between the stroke input and the output image, and assess realism using Kernel Inception Distance (KID) between real and generated images. 
Stroke inputs are generated from RGB images using the algorithm provided in SDEdit.
Given the subjective nature of image editing, we conduct a large-scale user study to calculate the user preference metric. 

For face editing, we evaluate identity preservation, prompt alignment, and overall image quality using a face recognition metric~\citep{ruiz2024hyperdreambooth}, CLIP-T scores~\citep{clip}, and  using CLIP-I scores~\citep{clip}, respectively.
For the face recognition score, we calculate the L2 distance between the face embedding of the original image and the edited image, obtained from Inception ResNet trained on CASIA-Webface dataset.
Similar to SDEdit~\citep{sdedit}, we conduct extensive experiments on Stroke2Image generation, and showcase additional capabilities qualitatively on a wide variety of semantic image editing tasks.

\textbf{Algorithm.}
The pseudo-code for getting the structured noise is provided in \textbf{Algorithm~\ref{alg:controlled-fwd-ode}}, and transforming that noise back to an image is given in \textbf{Algorithm~\ref{alg:controlled-rev-ode}}.

\begin{algorithm}[!t]
    \caption{Controlled Forward ODE \eqref{eq:controlled-ODE} }
    \label{alg:controlled-fwd-ode}
         \KwIn{Discretization steps $N$,
         reference image $\rvy_0$,
         prompt embedding network $\Phi$,
         Flux model $u(\cdot,\cdot, \cdot;\varphi)$,
         Flux noise scheduler $\sigma: [0,1]\rightarrow \R$\\
         \textbf{Tunable parameter:} Controller guidance $\gamma$
         }
        \KwOut{Structured noise $Y_1$}
        Initialize $Y_0  = \rvy_0$\\
        Fix a noise sample $\rvy_1$\\
        \For{$i=0$ \KwTo $N-1$}{
        Current time step: $t_i = \frac{i}{N}$\\
        Next time step: $t_{i+1} = \frac{i+1}{N}$\\
        Unconditional vector field: $u_{t_i}\left(Y_{t_i}\right) = u(Y_{t_i}, t_i,\Phi(\text{``"});\varphi)$ \hfill $\triangleright$ \textbf{Proposition}~\ref{prop:inv-wo-control}\\
        Conditional vector field: $u_{t_i}\left(Y_{t_i}|\rvy_1\right) = \frac{\rvy_1 - Y_{t_i}}{1-t_i} $ \hfill $\triangleright$ \textbf{Proposition}~\ref{prop:rf-lqr}\\
        Controlled vector field: $\hat{u}_{t_i}(Y_{t_i}) = u_{t_i}\left(Y_{t_i}\right) + \gamma \left(u_{t_i}\left(Y_{t_i}|\rvy_1\right) - u_{t_i}\left(Y_{t_i}\right)\right)$ \hfill $\triangleright$ODE~\eqref{eq:controlled-ODE}\\
        Next state: $Y_{t_{i+1}} = Y_{t_i} + \hat{u}_{t_i}(Y_{t_i})\left(\sigma(t_{i+1}) - \sigma(t_{i})\right)$
        }
        \Return $Y_1$
\end{algorithm}

\begin{algorithm}[!t]
    \caption{Controlled Reverse ODE \eqref{eq:gen-ode-w-controller}}
    \label{alg:controlled-rev-ode}
         \KwIn{Discretization steps $N$,
         reference text ``prompt",
         reference image $\rvy_0$,
         prompt embedding network $\Phi$,
         Flux model $u(\cdot,\cdot, \cdot;\varphi)$,
         Flux noise scheduler $\sigma: [0,1]\rightarrow \R$,\\
         \hspace{1.1cm}structured noise $\rvy_1$ \\
         \textbf{Tunable parameter:} Controller guidance $\eta$
         }
        \KwOut{Edited image $X_1$}
        Initialize $X_0  = \rvy_1$\\
        \For{$i=0$ \KwTo $N-1$}{
        Current time step: $t_i = \frac{i}{N}$\\
        Next time step: $t_{i+1} = \frac{i+1}{N}$\\
        Unconditional vector field: $v_{t_i}\left(X_{t_i}\right) = -u(X_{t_i}, 1-t_i,\Phi(\text{prompt});\varphi)$ \hfill $\triangleright$ \textbf{Proposition}~\ref{prop:inv-wo-control}\\
        Conditional vector field: $v_{t_i}\left(X_{t_i}|\rvy_0\right) = \frac{\rvy_0 - X_{t_i}}{1-t_i} $ \hfill $\triangleright$ \textbf{Proposition}~\ref{prop:rf-lqr}\\
        Controlled vector field: $\hat{v}_{t_i}(X_{t_i}) = v_{t_i}\left(X_{t_i}\right) + \eta \left(v_{t_i}\left(X_{t_i}|\rvy_0\right) - v_{t_i}\left(X_{t_i}\right)\right)$ \hfill $\triangleright$ODE~\eqref{eq:gen-ode-w-controller}\\
        Next state: $X_{t_{i+1}} = X_{t_i} + \hat{v}_{t_i}(X_{t_i})\left(\sigma(t_{i+1}) - \sigma(t_{i})\right)$
        }
        \Return $X_1$
\end{algorithm}

\subsection{Hyper-parameter configurations}
\label{sec:hparam-config}
In Table~\ref{tab:hparam}, we provide the hyper-parameters for the empirical results reported in \S\ref{sec:exps}. 
We use a fix $\gamma=0.5$ in our controlled forward ODE~\eqref{eq:controlled-ODE} and a time-varying guidance parameter $\eta_t$ in our controlled reverse ODE~\eqref{eq:gen-ode-w-controller}, as motivated in \textbf{Remark~\ref{rmk:controlled-ode}} and \textbf{Remark~\ref{rmk:controlled-sde}}.
Thus, our algorithm introduces one additional hyper-parameter $\eta_t$ into the Flux pipeline. 
For each experiment, we use a fixed time-varying schedule of $\eta_t$ described by starting time ($s$), stopping time $\tau$, and strength ($\eta$). 
We use the default config for Flux model: 3.5 for classifier-free guidance and 28 for the total number of inference steps.

\begin{table}[!tbh]
\centering
\caption{
Hyper-parameter configuration of our method for inversion and editing tasks. 
}
\label{tab:hparam}
\begin{tabular}{lccc}
\toprule
\textbf{Task} & \textbf{Starting Time ($s$)} & \multicolumn{2}{c}{\textbf{Controller Guidance ($\eta_t$)}} \\
\cmidrule(lr){3-4}
 &  & \textbf{Stopping Time ($\tau$)} & \textbf{Strength ($\eta$)} \\
\midrule
Stroke2Image  & 3 & 5 & 0.9 \\
Object insert & 0 & 6 & 1.0 \\
Gender editing  & 0 & 8 & 1.0 \\
Age editing & 0 & 5 & 1.0 \\
Adding glasses & 6 & 25 & 0.7\\
Stylization & 0 & 6 & 0.9 \\
\bottomrule
\end{tabular}
\end{table}

\subsection{Ablation Study}
\label{sec:ablation}
In this section, we conduct ablation study for our controller guidance parameter $\eta_t$.
We consider two different time-varying schedules for $\eta_t$, and show that our controller strength allows for a smooth interpolation between unconditional and conditional generation.

\begin{figure}[!tbh]
\includegraphics[width=\linewidth]{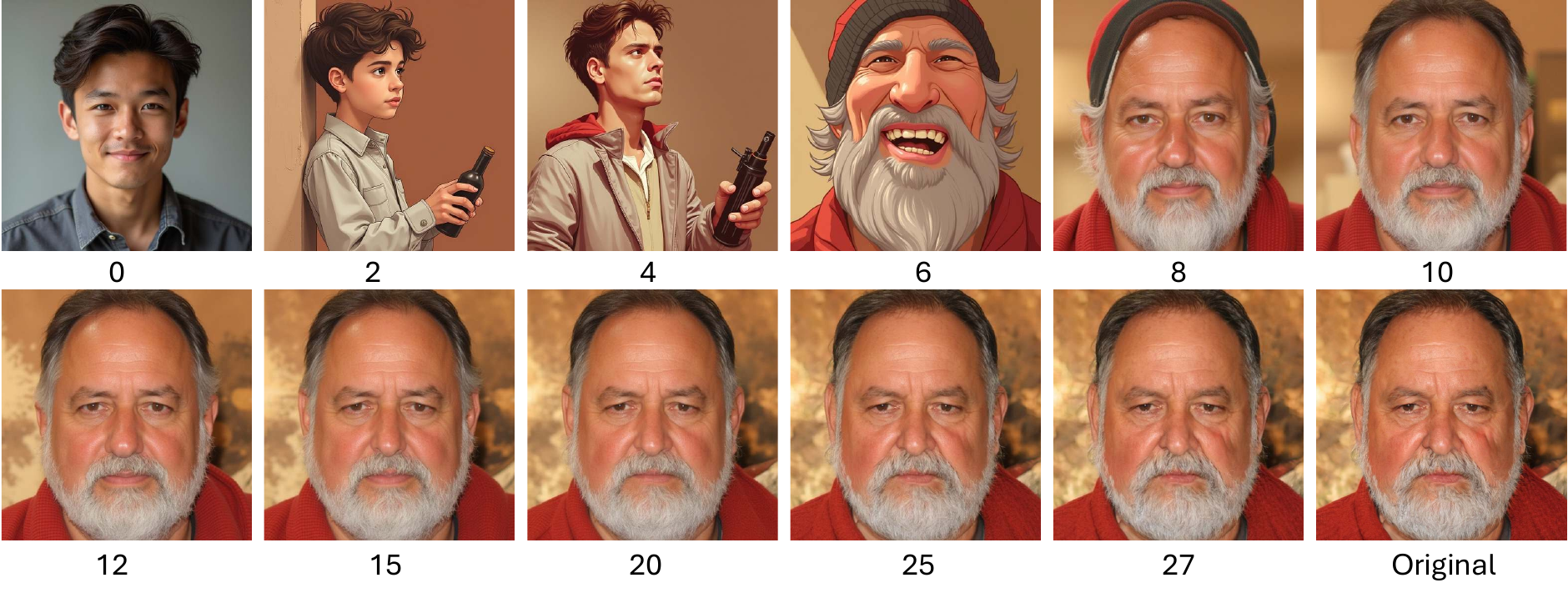}
\caption{
\textbf{Effect of starting time.} 
Prompt: ``A young man". 
The number below each figure denotes the starting time scaled by 28 (the total number of denoising steps) for better interpretation. 
In the absence of controller guidance ($\eta_t=0$), increasing the starting time ($s$) in our controlled ODE~\eqref{eq:gen-ode-w-controller} improves faithfulness to the original image.
}
\label{fig:abl-start-time}
\end{figure}

In \Figref{fig:abl-start-time}, we show the effect of starting time in controlling the faithfulness of inversion; starting time $s \in [0,1]$ is defined as the time at which our controlled reverse ODE~\eqref{eq:gen-ode-w-controller} is initialized. 
The initial state $X_s=\rvy_{1-s}$ is obtained by integrating the controlled forward ODE~\eqref{eq:controlled-ODE} from $0 \rightarrow 1-s$.

\begin{figure}[!tbh]
\includegraphics[width=\linewidth]{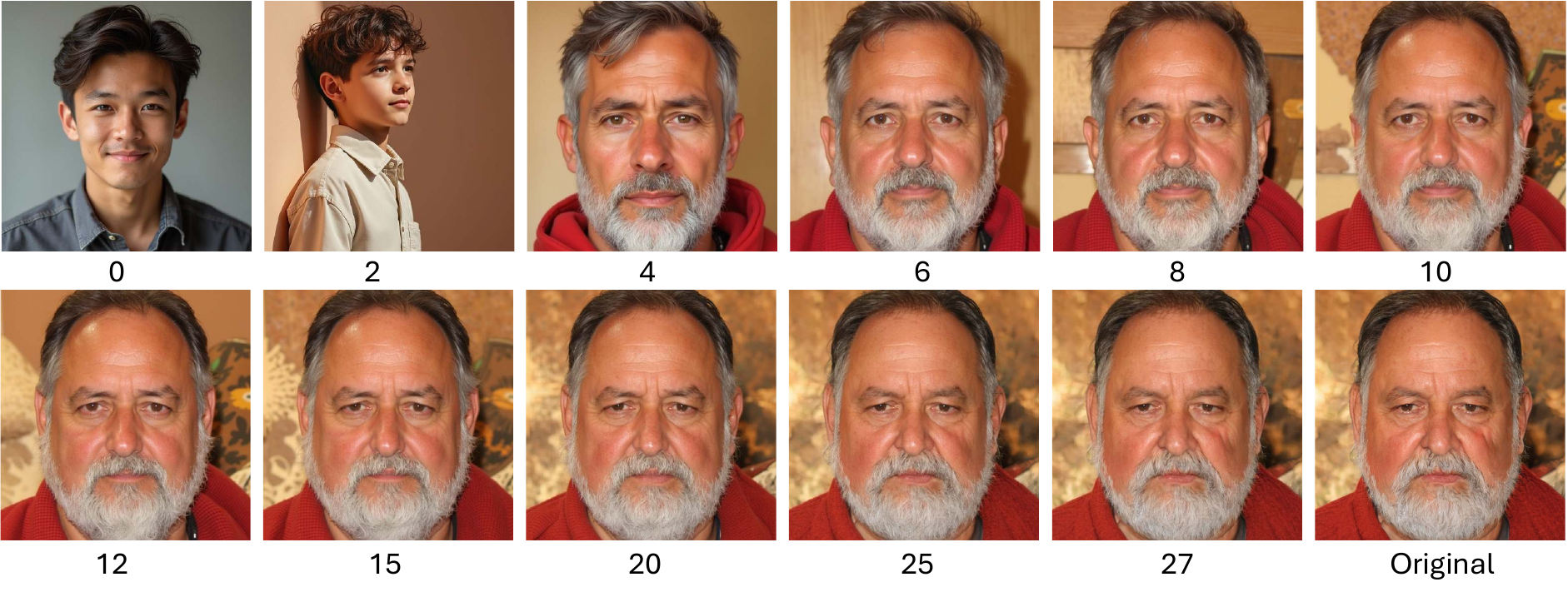}
\caption{
\textbf{Effect of controller guidance.}
Prompt: ``A young man". 
For a fixed starting time $s=0$, consider a time-varying controller guidance schedule $\eta_t = \eta ~\forall t \leq \tau$ and $0$ otherwise. 
The number below each figure denotes the stopping time $\tau$ scaled by 28 (the total number of denoising steps) for better interpretation.
Increasing $\tau$ increases the controller guidance ($\eta_t$) that improves faithfulness to the original image.
}
\label{fig:abl-stop-time}
\end{figure}

In \Figref{fig:abl-stop-time}, we study the effect of stopping time.
We find that increasing controller guidance $\eta_t$ by increasing the stopping time $\tau$ guides the reverse flow towards the original image. 
However, we observe a phase transition around $\tau = 0.14 = 4/28 $, indicating that the resulting drift in our controlled reverse ODE~\eqref{eq:gen-ode-w-controller} is dominated by the conditional vector field $v_t(X_t|\rvy_0)$ for $t\geq \tau$.
Therefore, the reverse flow solves the LQR problem~\eqref{eq:gen-lqr-rev} and drives toward the terminal state (i.e., the original image).

\begin{figure}[!tbh]
\includegraphics[width=\linewidth]{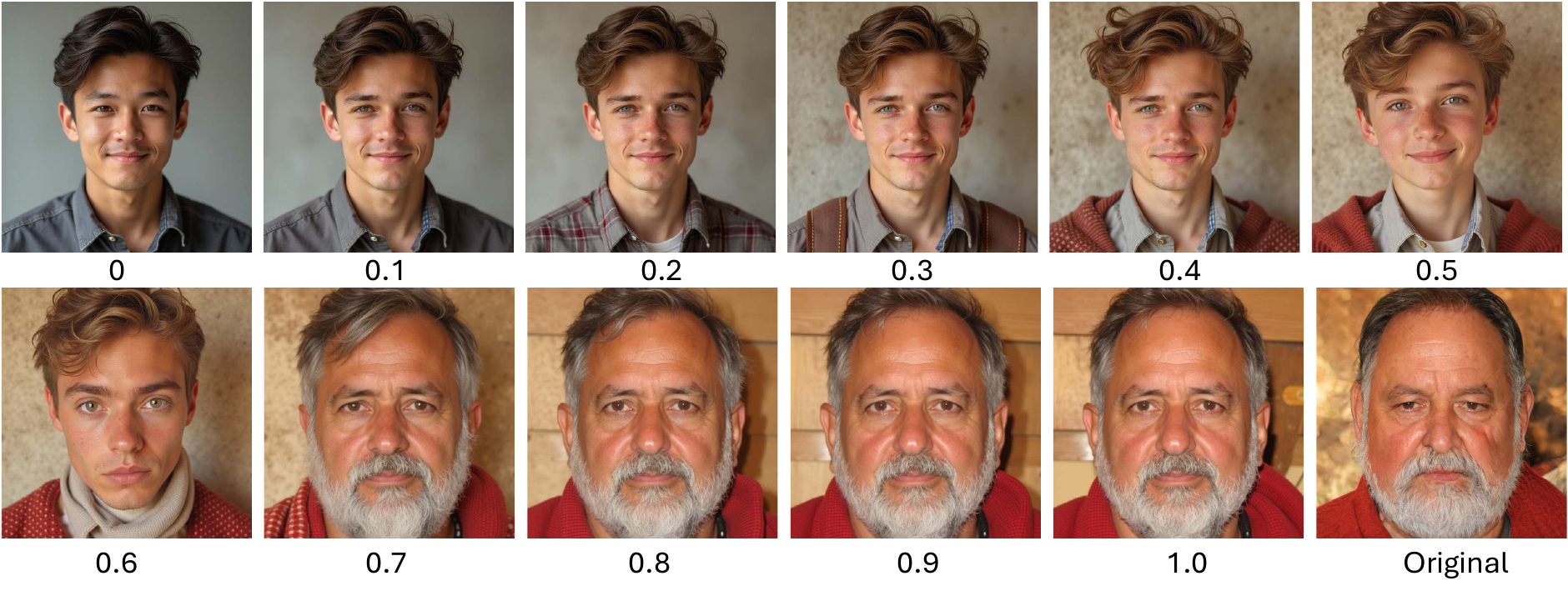}
\caption{
\textbf{Effect of controller guidance} for another time-varying schedule. 
Prompt: ``A young man". 
The number below each figure denotes the starting time scaled by 28 (the total number of denoising steps) for better interpretation.
For a fixed starting time $s=0$ and stopping time $\tau=8$, consider a time-varying controller guidance schedule $\eta_t = \eta ~\forall t \leq \tau$ and $0$ otherwise. 
Increasing $\eta$ increases the controller guidance ($\eta_t$) that improves faithfulness to the original image.
}
\label{fig:control-strength-all}
\end{figure}

In \Figref{fig:control-strength-all}, we visualize the effect of our controller guidance for another time-varying schedule.
We make a similar observation as in \Figref{fig:abl-stop-time}: increasing $\eta_t$ improves faithfulness. 
However, we notice a smooth transition from the unconditional to the conditional vector field, evidence from the smooth interpolation between ``A young man" at the top left ($\eta=0$) and the original image at the bottom right.

\subsection{Numerical Simulation}
\label{sec:num_sim}
In this section, we design synthetic experiments to compare reconstruction accuracy of DM and RF inversion.
Given $Y_0 \sim p_0$, where the data distribution $p_0\coloneqq \gN(\mu,I)$ and the source distribution $q_0 \coloneqq \gN(0,I)$, we numerically simulate the ODEs and SDEs associated with DM and RF inversion; see our discussion in \S\ref{sec:method}.

For $\mu=10$, we fix $\gamma=0.5$ in the controlled forward ODE~\eqref{eq:controlled-ODE}, and $\eta=0.5$ in the controlled reverse ODE~\eqref{eq:gen-ode-w-controller}.
These ODEs are simulated using the Euler discretization scheme with 100 steps. 
Additionally, we simulate the uncontrolled rectified flow ODEs~\eqref{eq:inv-wo-control} $\rightarrow$ \eqref{eq:gen-ode} as a special case of our controlled ODEs~\eqref{eq:controlled-ODE} $\rightarrow$ \eqref{eq:gen-ode-w-controller} by setting $\gamma=\eta=0$, and the deterministic diffusion model DDIM~\citep{ddim} in the same experimental setup. 

The inversion accuracy is reported in Table~\ref{tab:comp-faith-gauss}.
Observe that RF inversion has less L2 and L1 error compared to DDIM inversion~\eqref{eq:ddim}.
The minimum error is obtained by setting $\gamma=\eta=0$ (i.e., reversing the standard rectified flows), which supports our discussion in \S\ref{sec:rf-inv}. 

Furthermore, we simulate the stochastic samplers corresponding to these ODEs in Table~\ref{tab:comp-faith-gauss}, highlighted in {\color{orange}orange}.
Similar to the deterministic samplers, we observe that stochastic equivalents of rectified flows more accurately recover the original sample compared to diffusion models.
Our controller in RF Inversion \eqref{eq:sde-controlled-vector-field} $\rightarrow$ \eqref{eq:stoch-rect-flow-cond-sampling} effectively reduces the reconstruction error in the uncontrolled RF Inversion \eqref{eq:sde-optimal-vector-field} $\rightarrow$ \eqref{eq:stoch-rect-flow-sampling}, which are special cases when $\gamma=\eta=0$.
Thus, we demonstrate that (controlled) rectified stochastic processes are better at inverting a given sample from the target distribution, outperforming the typical OU process used in diffusion models~\citep{ncsn,ddpm,ddim,songscore}.

\begin{table}[!tbh]
\caption{DM and RF inversion accuracy. Stochastic samplers are highlighted in {\color{orange}orange}.}  
\label{tab:comp-faith-gauss}
\centering
\begin{tabular}{lcr}
\toprule
Method & L2 Error & L1 Error \\
\midrule
DDIM Inversion \eqref{eq:ddim} & 6.024 & 19.038 \\ 
\rowcolor{orange!25}
DDPM Inversion \eqref{eq:ddpm} & 6.007 & 15.758 \\ 
RF Inversion ($\gamma=\eta=0$) \eqref{eq:controlled-ODE} $\rightarrow$ \eqref{eq:gen-ode-w-controller} & 0.092 & 0.20 \\ 
\rowcolor{orange!25}
RF Inversion ($\gamma=\eta=0$) \eqref{eq:sde-controlled-vector-field} $\rightarrow$ \eqref{eq:stoch-rect-flow-cond-sampling} & 3.564 & 8.795 \\ 
RF Inversion ($\gamma=0.5, \eta=0$) \eqref{eq:controlled-ODE} $\rightarrow$ \eqref{eq:gen-ode-w-controller} & 4.777 & 11.628 \\ 
RF Inversion ($\gamma=0, \eta=0.5$) \eqref{eq:controlled-ODE} $\rightarrow$ \eqref{eq:gen-ode-w-controller} & 1.219 & 3.074 \\ 
RF Inversion ($\gamma=0.5, \eta=0.5$) \eqref{eq:controlled-ODE} $\rightarrow$ \eqref{eq:gen-ode-w-controller} & 0.628 & 1.643 \\ 
\rowcolor{orange!25}
RF Inversion ($\gamma=\eta=0.5$) \eqref{eq:sde-controlled-vector-field} $\rightarrow$ \eqref{eq:stoch-rect-flow-cond-sampling} & 0.269 & 0.694 \\ 
\rowcolor{orange!25}
RF Inversion ($\gamma=\eta=1.0$) \eqref{eq:sde-controlled-vector-field} $\rightarrow$ \eqref{eq:stoch-rect-flow-cond-sampling} & 0.003 & 0.010 \\ 
\bottomrule
\end{tabular}
\end{table}

\begin{figure}[!tbh]
     \centering
     \begin{subfigure}[b]{0.24\columnwidth}
         \centering
         \includegraphics[width=\linewidth]{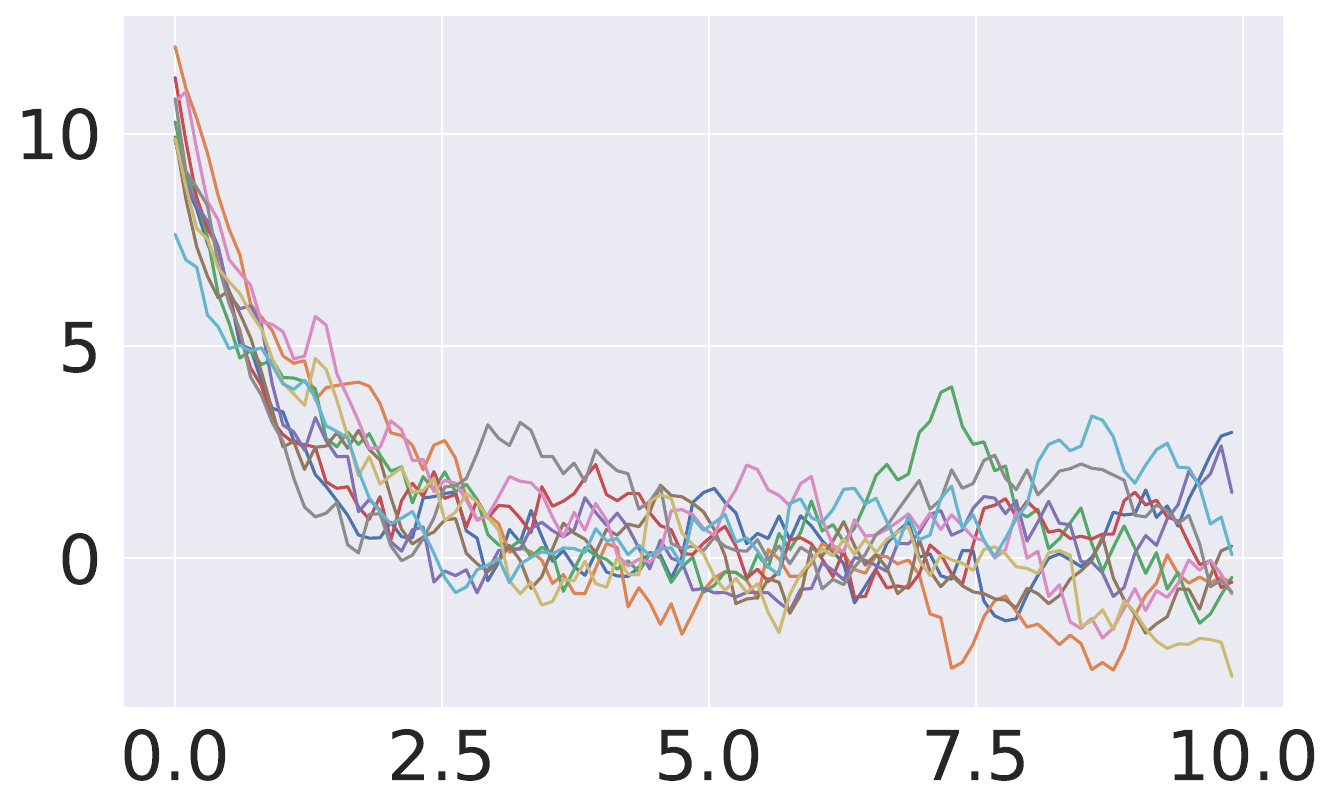}
        \caption{DDPM \eqref{eq:ddpm} Fwd.}
     \end{subfigure}
     \begin{subfigure}[b]{0.24\columnwidth}
         \centering
         \includegraphics[width=\linewidth]{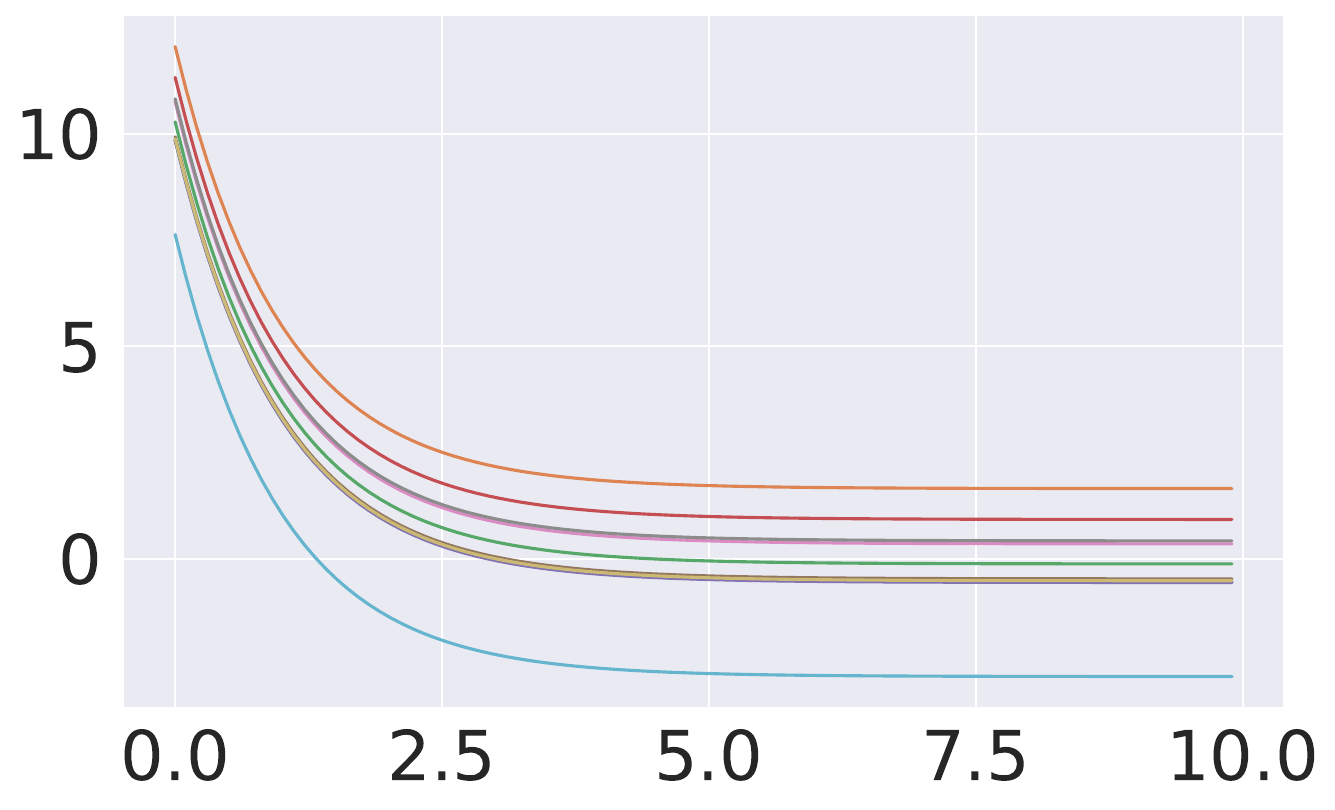}
        \caption{DDIM \eqref{eq:ddim} Fwd.}
     \end{subfigure}
     \begin{subfigure}[b]{0.24\columnwidth}
         \centering
         \includegraphics[width=\linewidth]{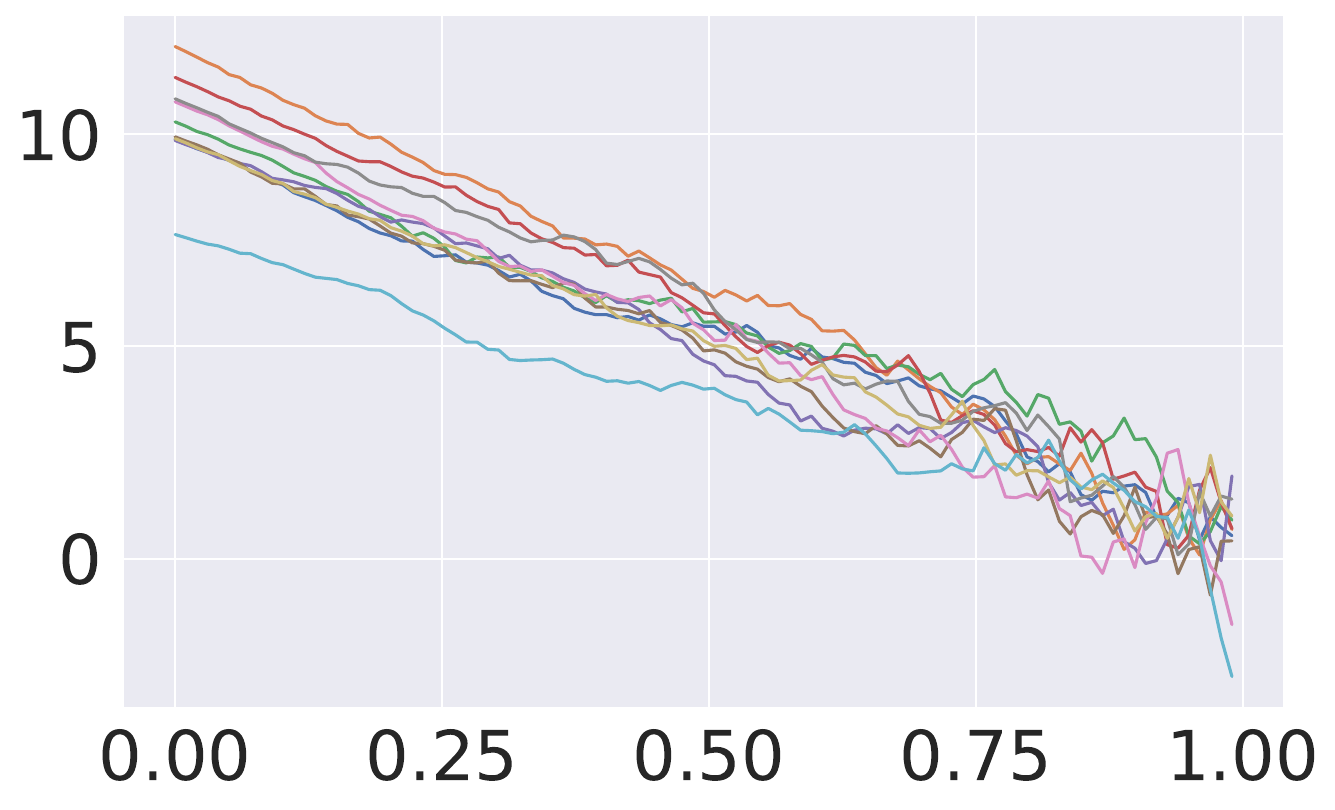}
        \caption{SDE~\eqref{eq:sde-optimal-vector-field} Fwd.}
     \end{subfigure}
     \begin{subfigure}[b]{0.24\columnwidth}
         \centering
         \includegraphics[width=\linewidth]{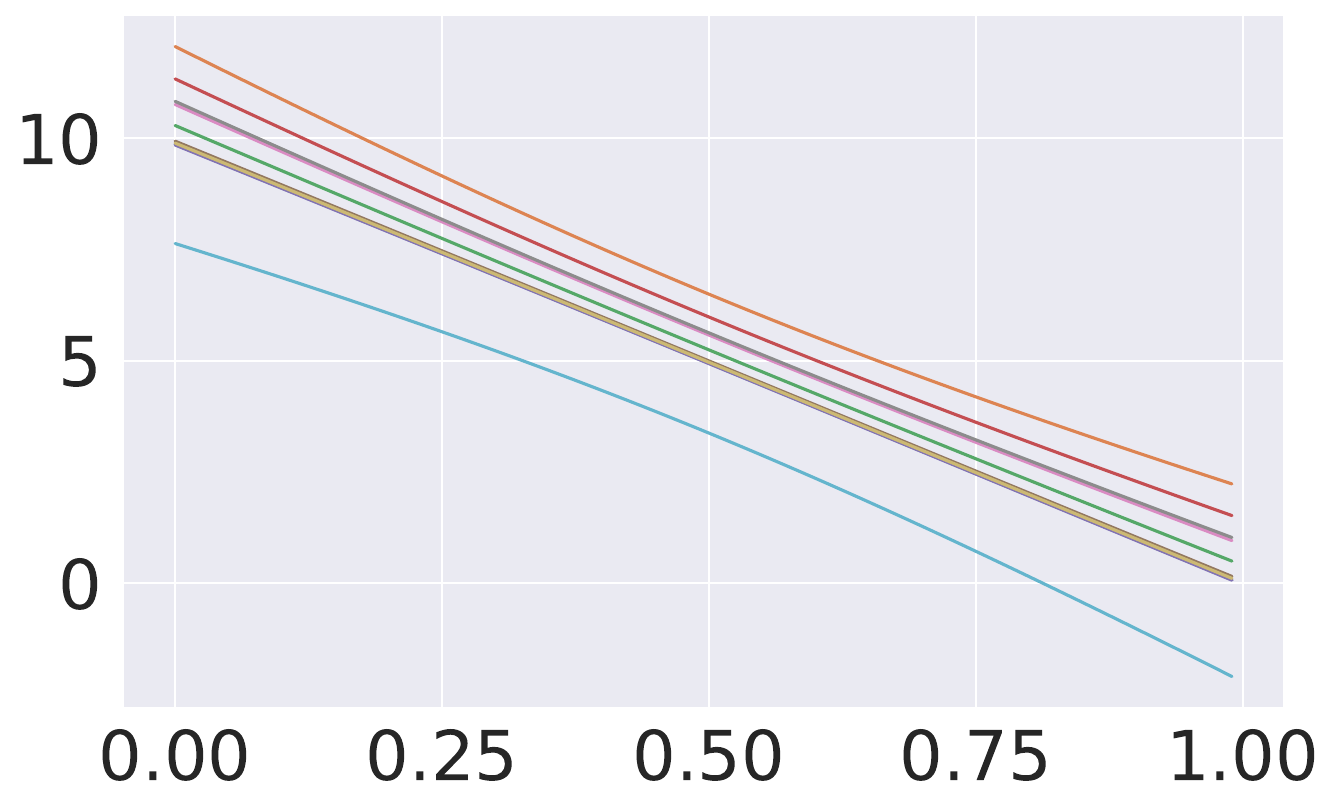}
        \caption{RF \eqref{eq:inv-wo-control} Fwd.}
     \end{subfigure}
    \\
     \begin{subfigure}[b]{0.24\columnwidth}
         \centering
         \includegraphics[width=\linewidth]{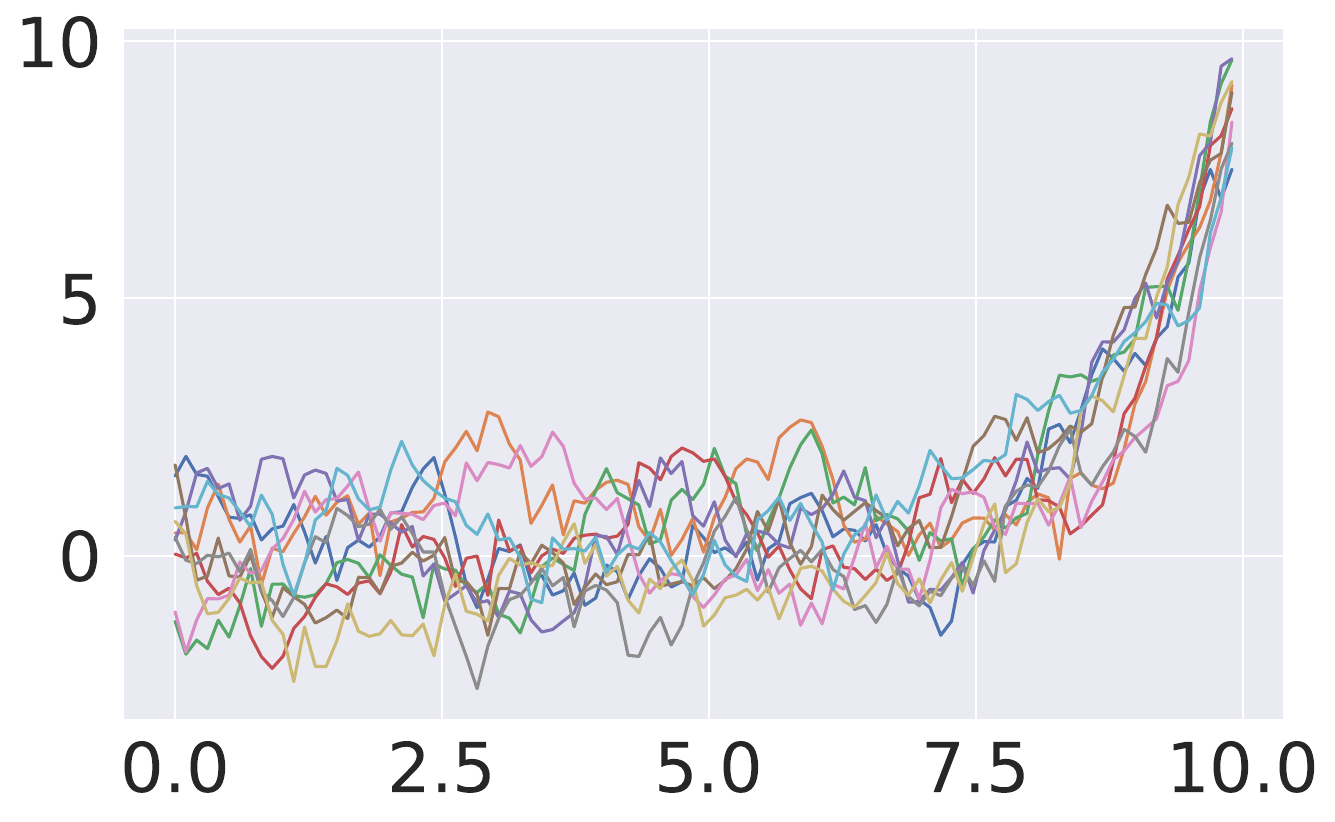}
        \caption{DDPM \eqref{eq:ddpm} Rev.}
     \end{subfigure}
     \begin{subfigure}[b]{0.24\columnwidth}
         \centering
         \includegraphics[width=\linewidth]{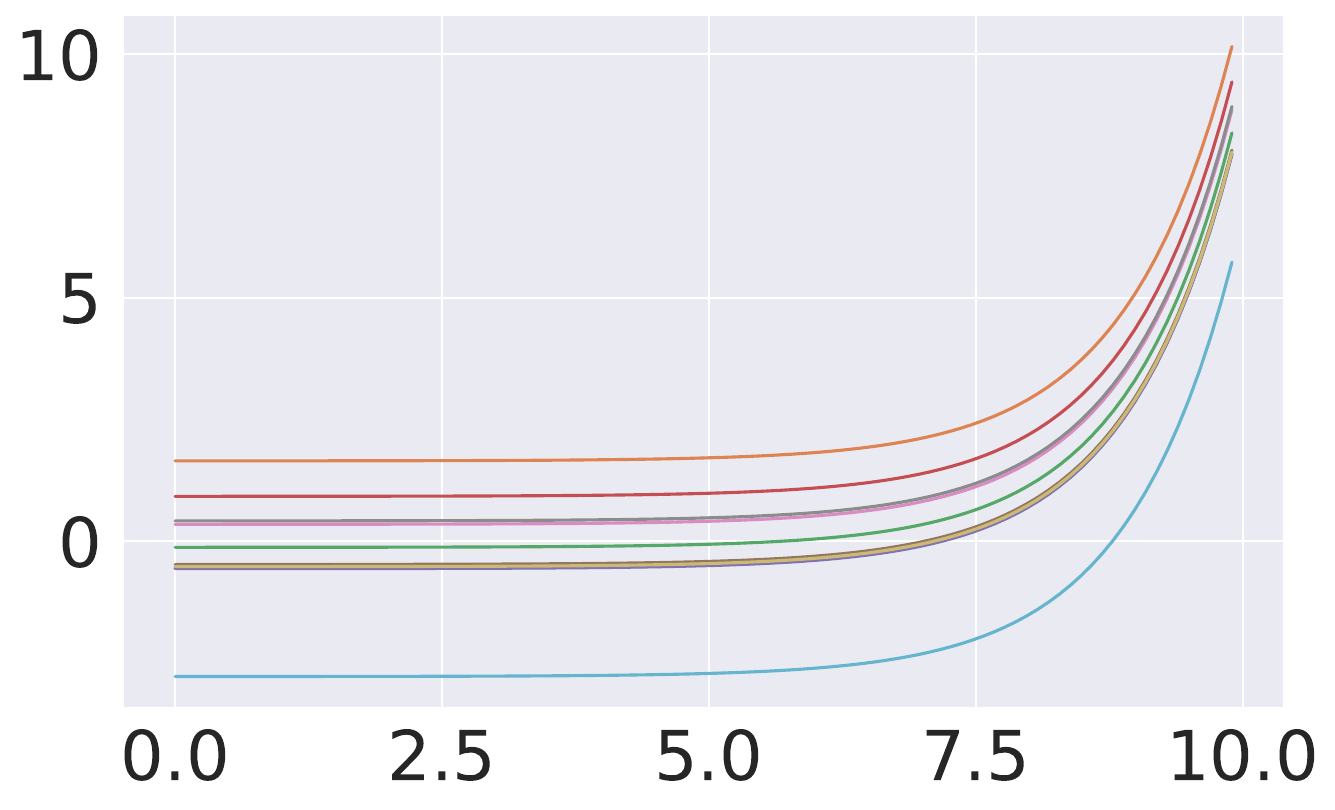}
        \caption{DDIM \eqref{eq:ddim} Rev.}
     \end{subfigure}
     \begin{subfigure}[b]{0.24\columnwidth}
         \centering
         \includegraphics[width=\linewidth]{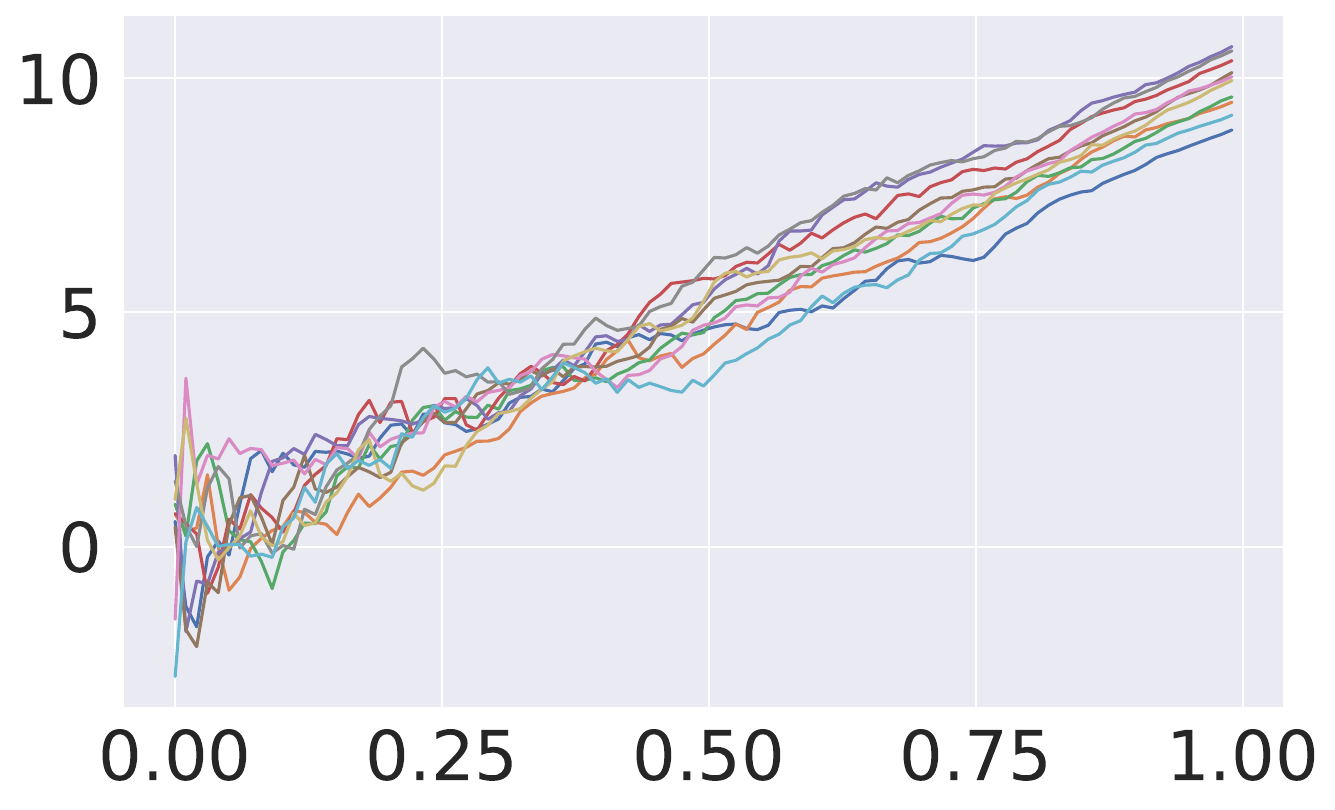}
        \caption{SDE \eqref{eq:sde-optimal-vector-field} Rev.}
     \end{subfigure}
     \begin{subfigure}[b]{0.24\columnwidth}
         \centering
         \includegraphics[width=\linewidth]{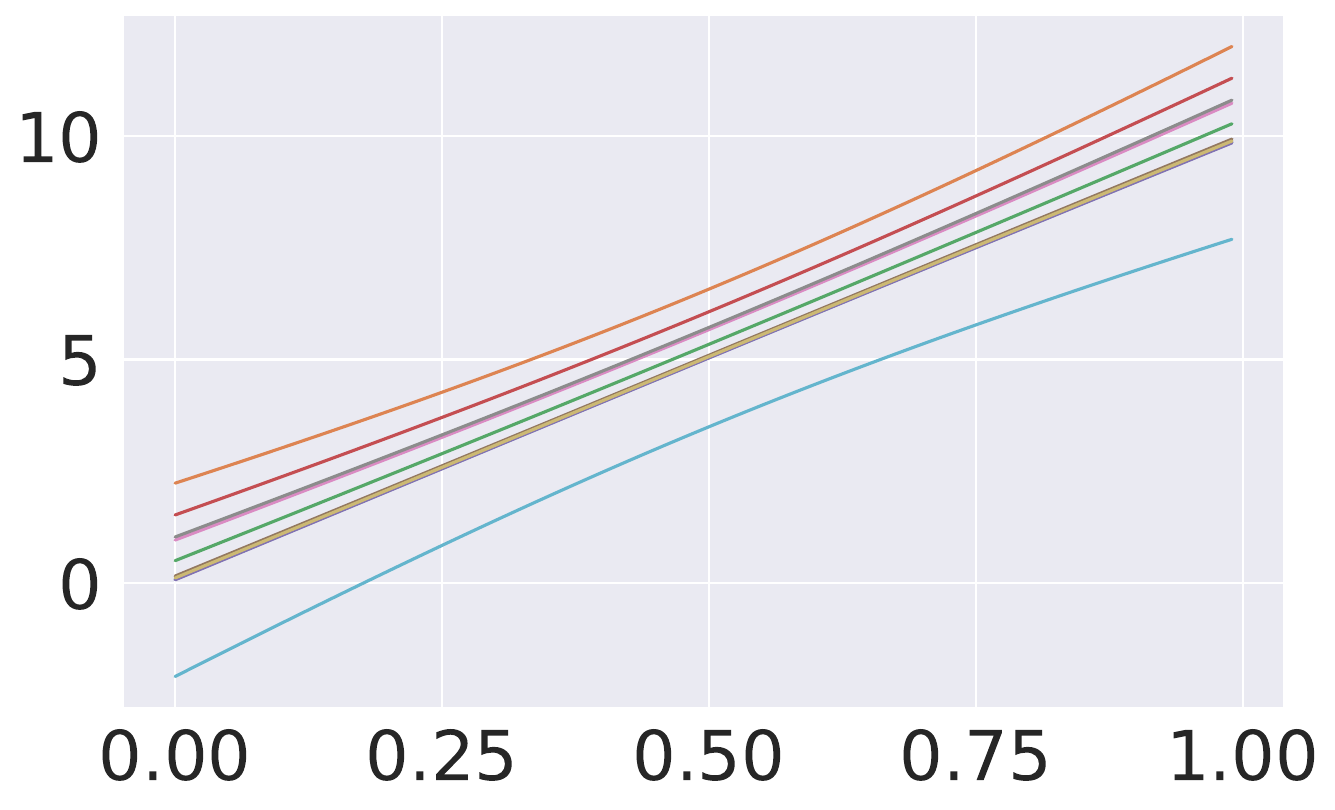}
        \caption{RF \eqref{eq:inv-wo-control} Rev.}
     \end{subfigure}
     \caption{
     \textbf{Sample paths of DMs and RFs.}
     Top row corresponds to the forward process $\{Y_t\}$, and bottom row, reverse process $\{X_t\}$.
     In each plot, time is along the horizontal axis and the process, along the vertical axis.
     The sample paths of RFs are straighter than that of DMs, allowing coarse discretization and faithful reconstruction. 
     }
    \label{fig:comp-dm-rf}
\end{figure}

\begin{figure}[!tbh]
     \centering
     \begin{subfigure}[b]{0.24\columnwidth}
         \centering
         \includegraphics[width=\linewidth]{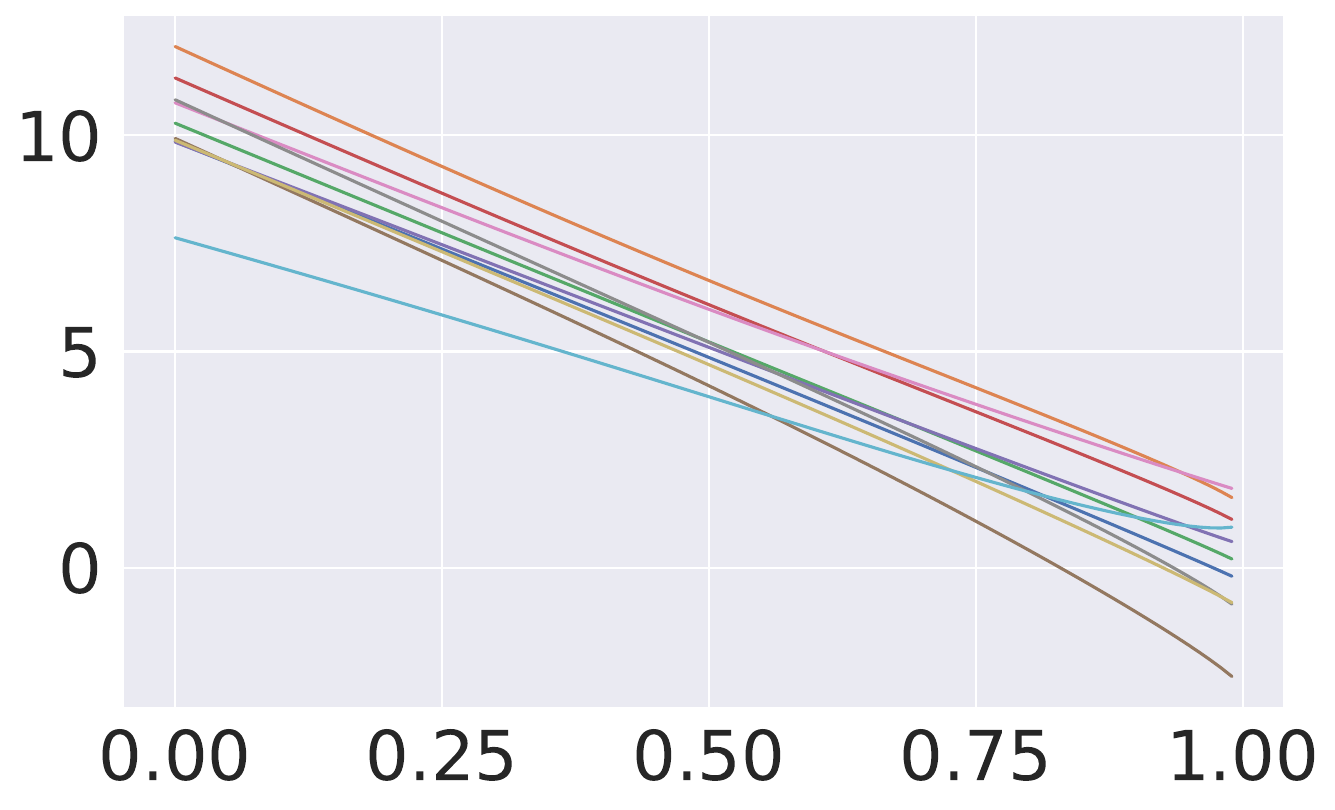}
        \caption{ODE \eqref{eq:controlled-ODE} Fwd.}
     \end{subfigure}
     \begin{subfigure}[b]{0.24\columnwidth}
         \centering
         \includegraphics[width=\linewidth]{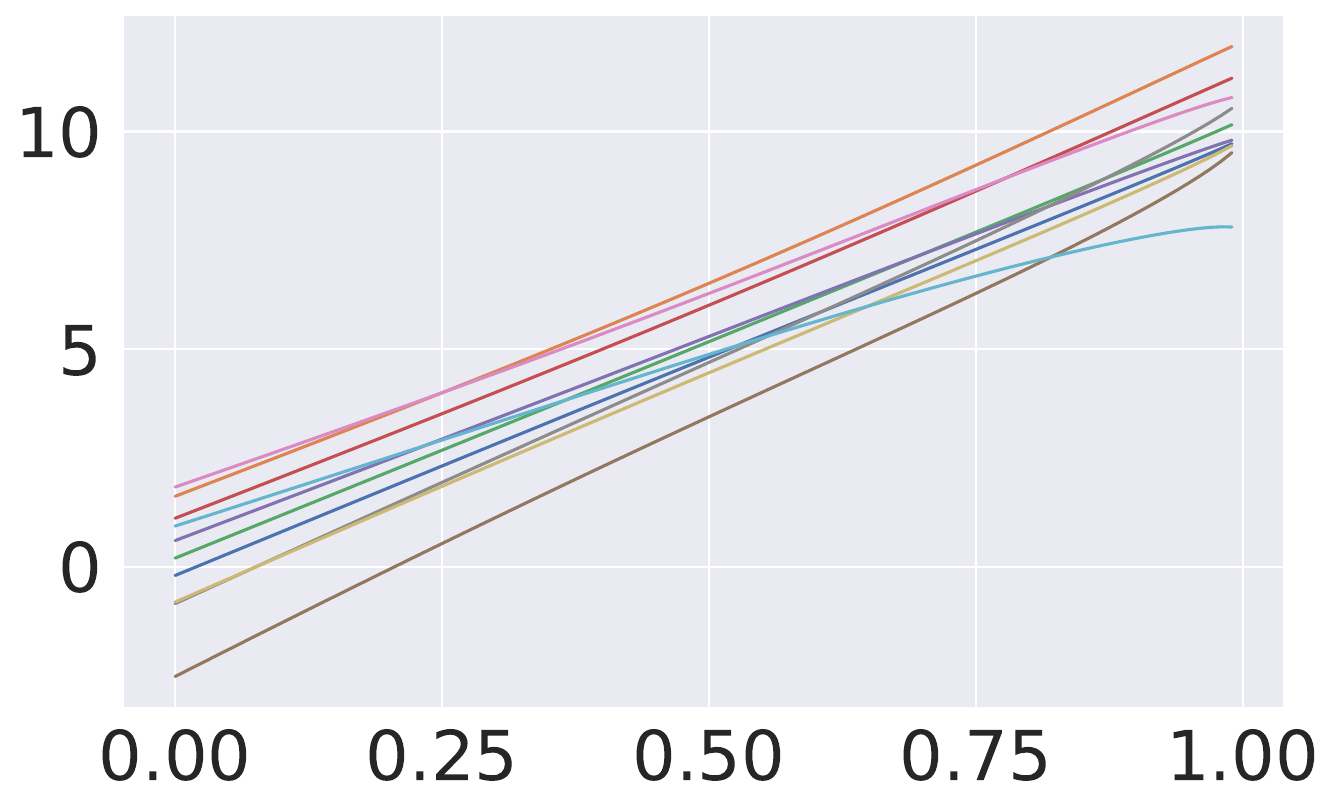}
        \caption{ODE \eqref{eq:gen-ode-w-controller} Rev.}
     \end{subfigure}
     \begin{subfigure}[b]{0.24\columnwidth}
         \centering
         \includegraphics[width=\linewidth]{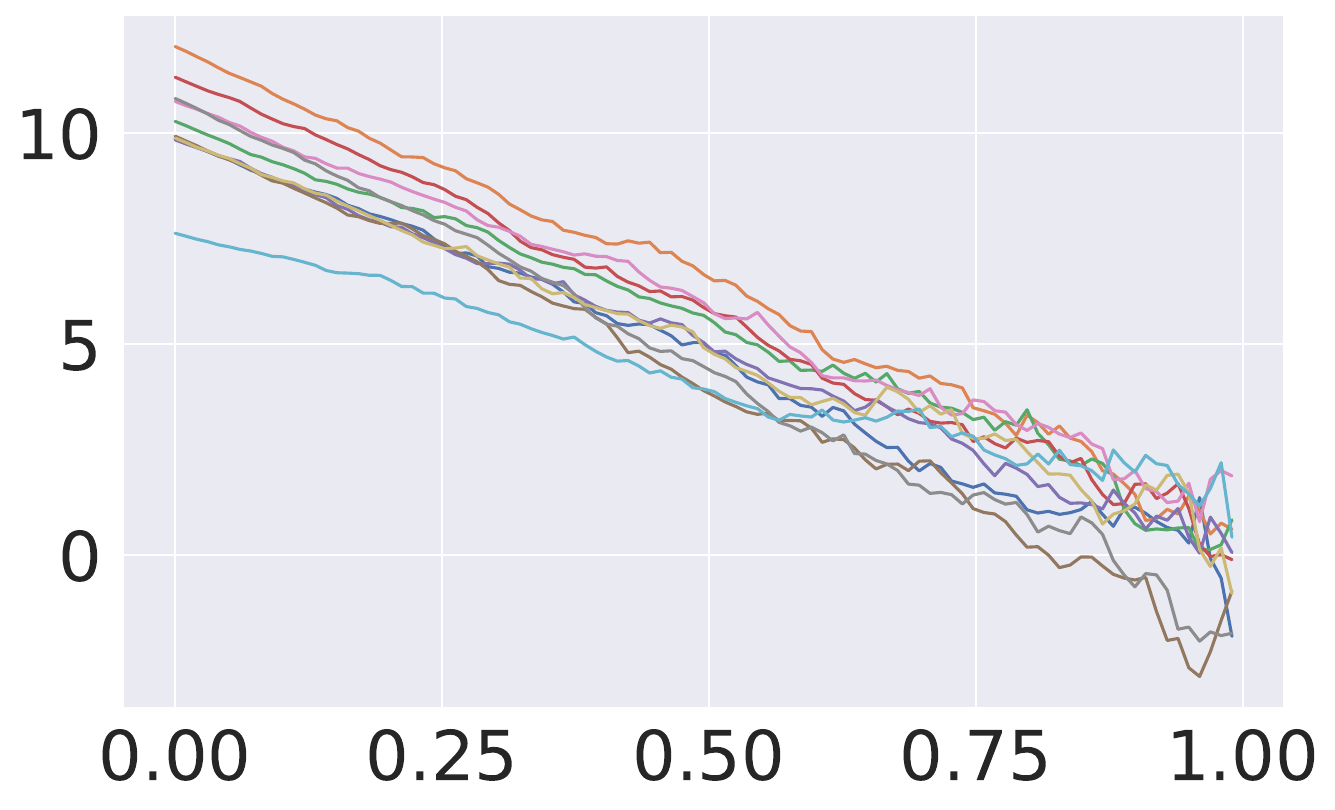}
        \caption{SDE \eqref{eq:sde-controlled-vector-field} Fwd.}
     \end{subfigure}
     \begin{subfigure}[b]{0.24\columnwidth}
         \centering
         \includegraphics[width=\linewidth]{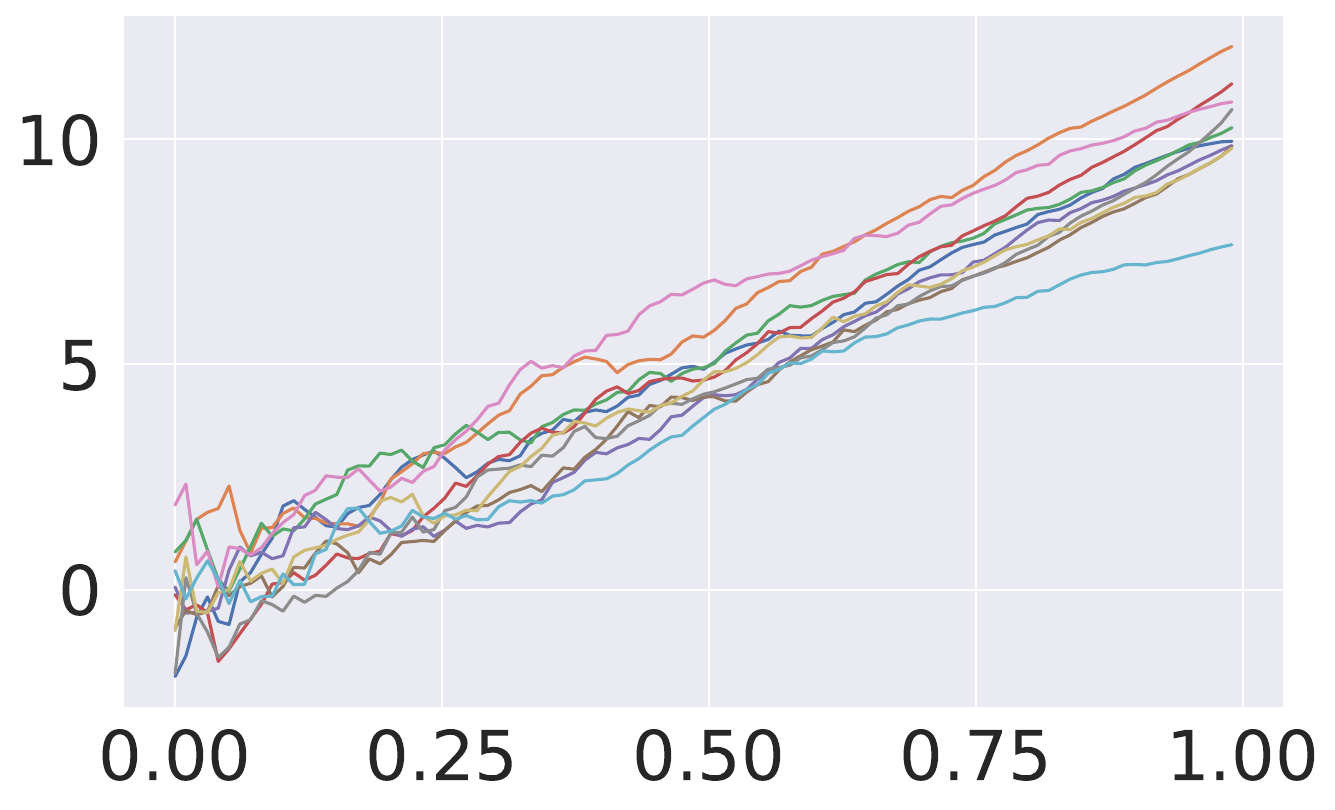}
        \caption{SDE \eqref{eq:stoch-rect-flow-cond-sampling} Rev.}
     \end{subfigure}
     \vspace{-0.2cm}
     \caption{
     \textbf{Sample paths of our controlled ODEs and SDEs.} 
     (a,c) The optimal controller $u_t(Y_t|Y_1)$ steers $Y_t$ towards the terminal state $Y_1\sim p_1$ during inversion.
     (b,d) Similarly, $v_t(X_t|Y_0)$ guides $X_t$ towards the reference image $Y_0 \sim p_0$, significantly reducing the reconstruction error. 
     }
    \label{fig:comp-crf-ode-sde}
\end{figure}

In \Figref{fig:comp-dm-rf}, we compare sample paths of diffusion models and recitified flows using 10 IID samples drawn from $p_0$.
In \Figref{fig:comp-crf-ode-sde}, we visualize paths for those samples using our controlled ODEs and SDEs with $\gamma= \eta=0.5$.

\subsection{Additional Results on Stroke2Image Generation}
\label{sec:addn-exp-results}
\begin{figure}[!tbh]
\includegraphics[width=\linewidth]{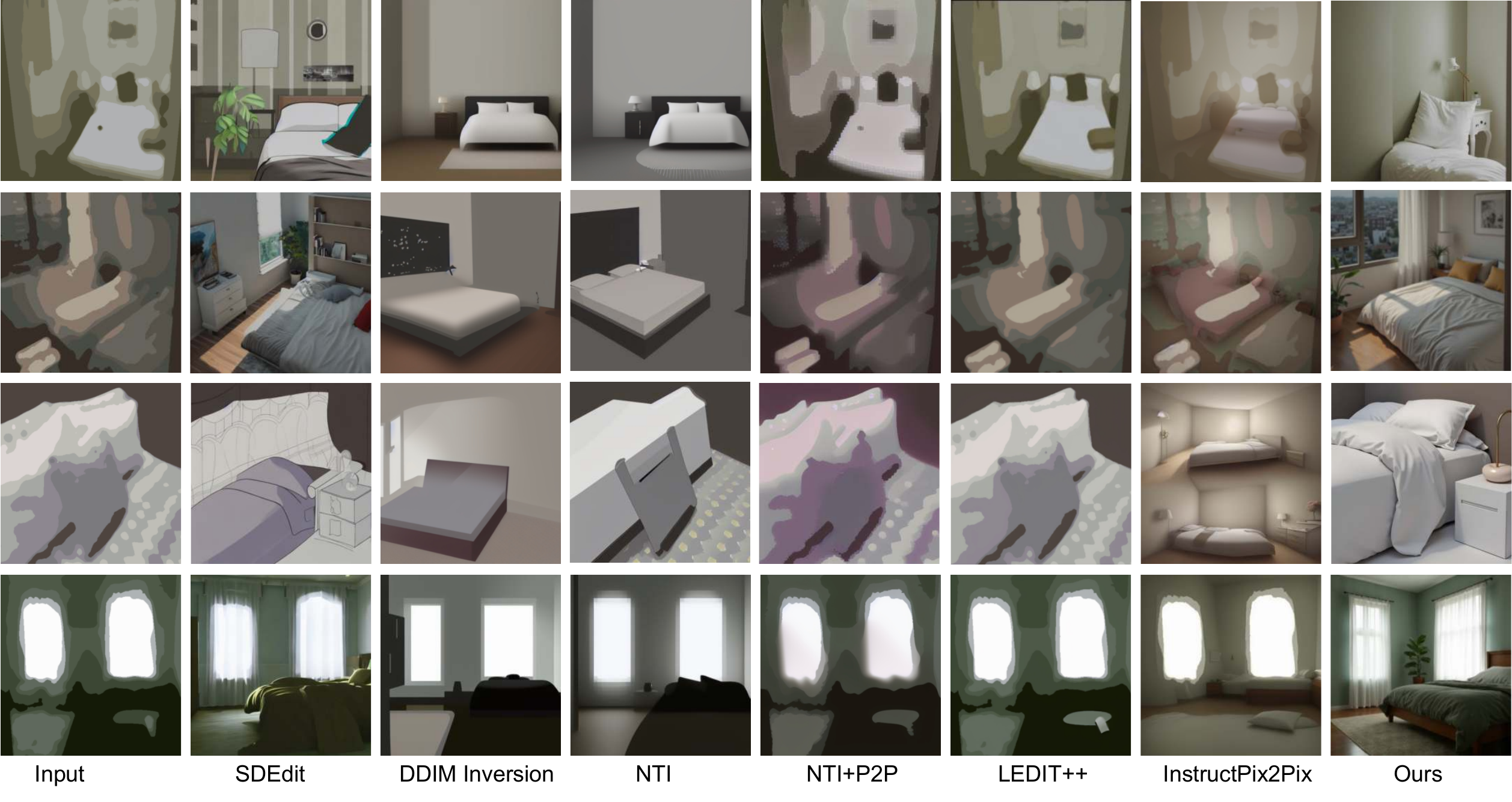}
\caption{
\textbf{Stroke2Image generation.}
Additional qualitative results on LSUN-Bedroom dataset comparing our method with SoTA training-free and training-based editing approaches. 
}
\label{fig:stroke2image-supp}
\end{figure}

\begin{figure}[!tbh]
\includegraphics[width=\linewidth]{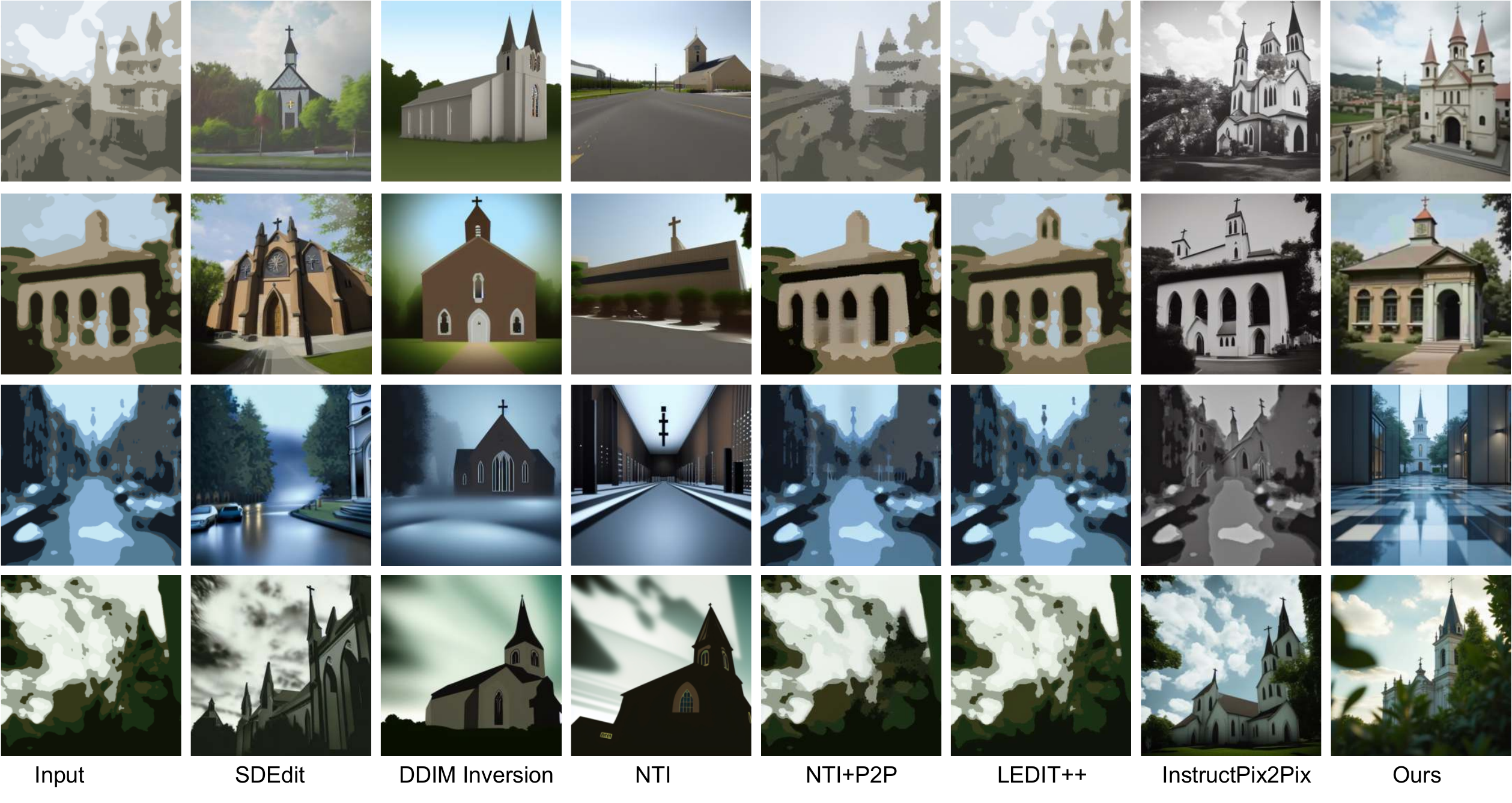}
\caption{
\textbf{Stroke2Image generation.}
Additional qualitative results on LSUN-Church dataset comparing our method with SoTA training-free and training-based editing approaches. 
}
\label{fig:stroke2image-church-supp}
\end{figure}

In \Figref{fig:stroke2image-supp} and \Figref{fig:stroke2image-church-supp}, we show additional qualitative results on Stroke2Image generation.
Our method generates more realistic images compared to leading training-free approaches in semantic image editing including optimization-based NTI~\citep{nti} and attention-based NTI+P2P~\citep{p2p}.
Furthermore, it gives a competitive advantage over the training-based approach InstructPix2Pix~\citep{instructpix2pix}.

\begin{figure}[!tbh]
\includegraphics[width=0.8\linewidth]{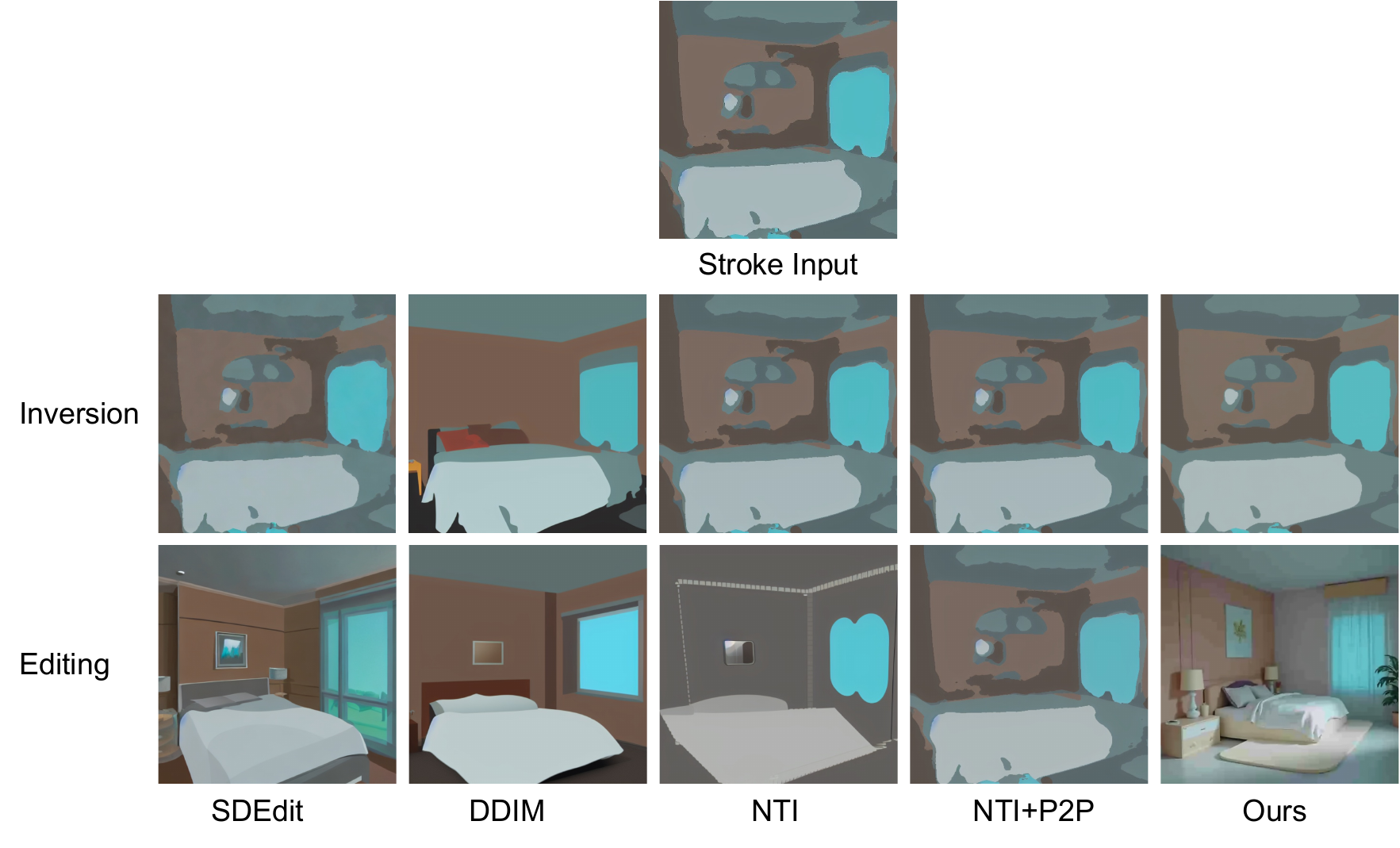}
\caption{
\textbf{Robustness.} 
For inversion, all methods perform well at recovering the stroke input when given a null prompt. However, when a new prompt like ``a photo-realistic picture of a bedroom" is provided, only our method successfully generates realistic images. The other methods continue to suffer from the initial corruption, failing to make the output more realistic.
}
\label{fig:faith-edit}
\end{figure}

In \Figref{fig:faith-edit}, we demonstrate the robustness of our approach to corruption at initialization.
All the methods transform the stroke input (corrupt image) to a structured noise, which is again transformed back to a similar looking stroke input, highlighting the faithfulness of these methods.
However, unlike our approach, the resulting images in other methods are not editable given a new prompt.

\subsection{Additional Results on Semantic Image Editing}
\label{sec:addn-exp-edit}

\begin{figure}[!tbh]
\includegraphics[width=\linewidth]{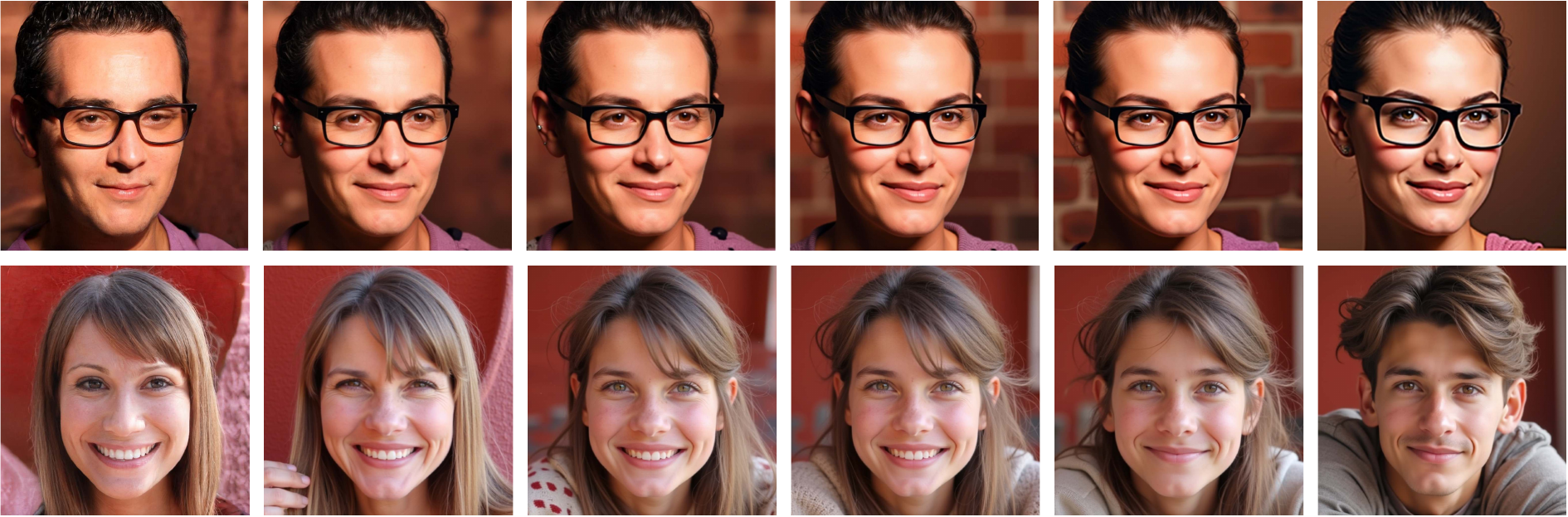}
\caption{
\textbf{Gender editing.} 
Our method smoothly interpolates between ``A man" $\leftrightarrow$ ``A woman".
}
\label{fig:gender-edit}
\end{figure}

\Figref{fig:gender-edit} illustrates a smooth interpolation between ``A man" $\rightarrow$ ``A woman" (top row) and ``A woman" $\rightarrow$ ``A man" (bottom row). The facial expression and the hair style are gradually morphed from one person to the other.

\begin{figure}[!tbh]
\includegraphics[width=\linewidth]{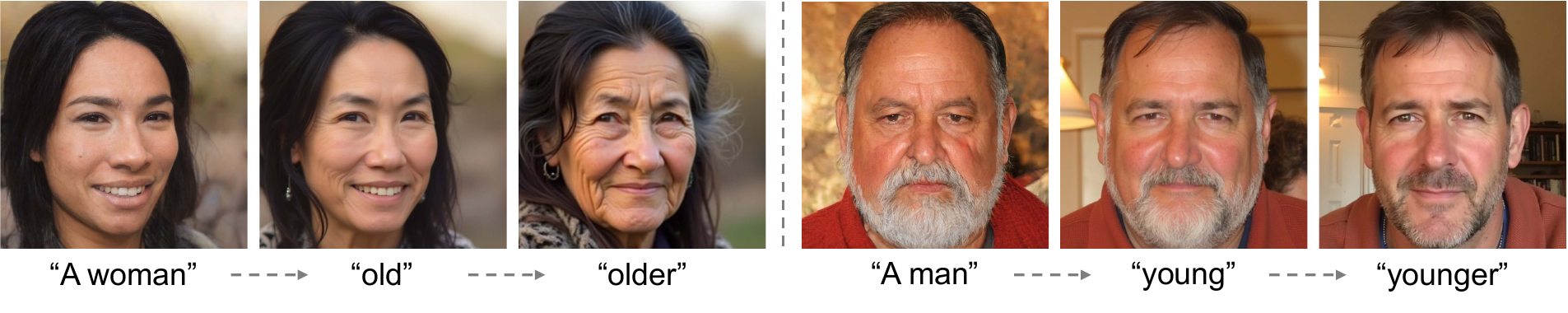}
\caption{
\textbf{Age editing.} 
Our method regulates the extent of age editing. 
}
\label{fig:age-edit}
\end{figure}

In \Figref{fig:age-edit}, we show the ability to regulate the extent of age editing. 
Given an image of a young woman and the prompt ``An old woman", we gradually reduce the controller strength $\eta_t$ to make the person look older. Similarly, we reduce the strength to make an old man look younger. 

\begin{figure}[!tbh]
\includegraphics[width=\linewidth]{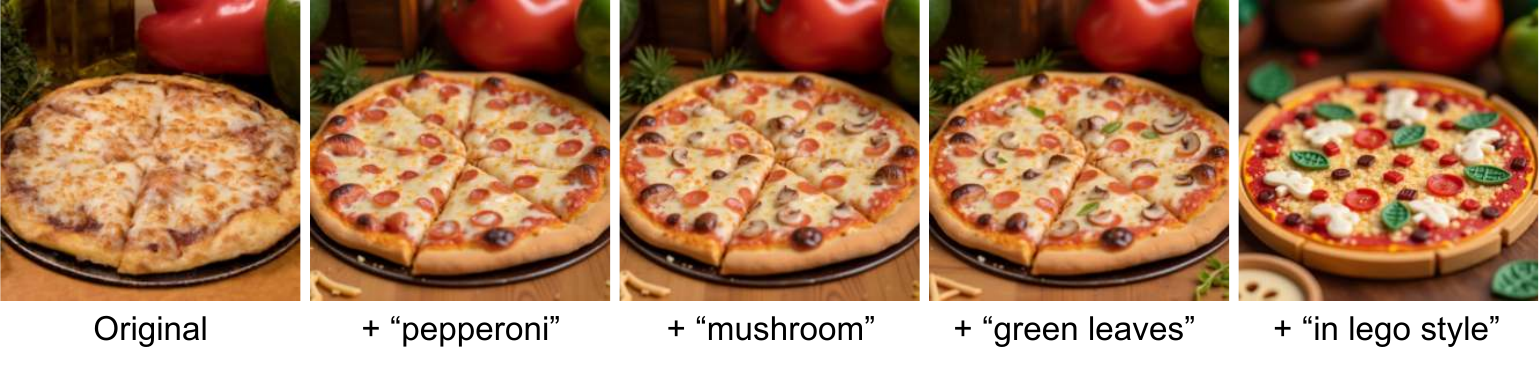}
\caption{
\textbf{Object insert.} 
Text-guided insertion of multiple objects sequentially.
}
\label{fig:seq-edit}
\end{figure}

\Figref{fig:seq-edit} shows the insertion of multiple objects by text prompts, such as ``pepperoni", ``mushroom", and ``green leaves" to an image of a pizza.
Interestingly, pepperoni is not deleted while inserting mushroom, and mushroom is not deleted while inserting green leaves.
The product is finally presented in a lego style.

\begin{figure}[!tbh]
\includegraphics[width=\linewidth]{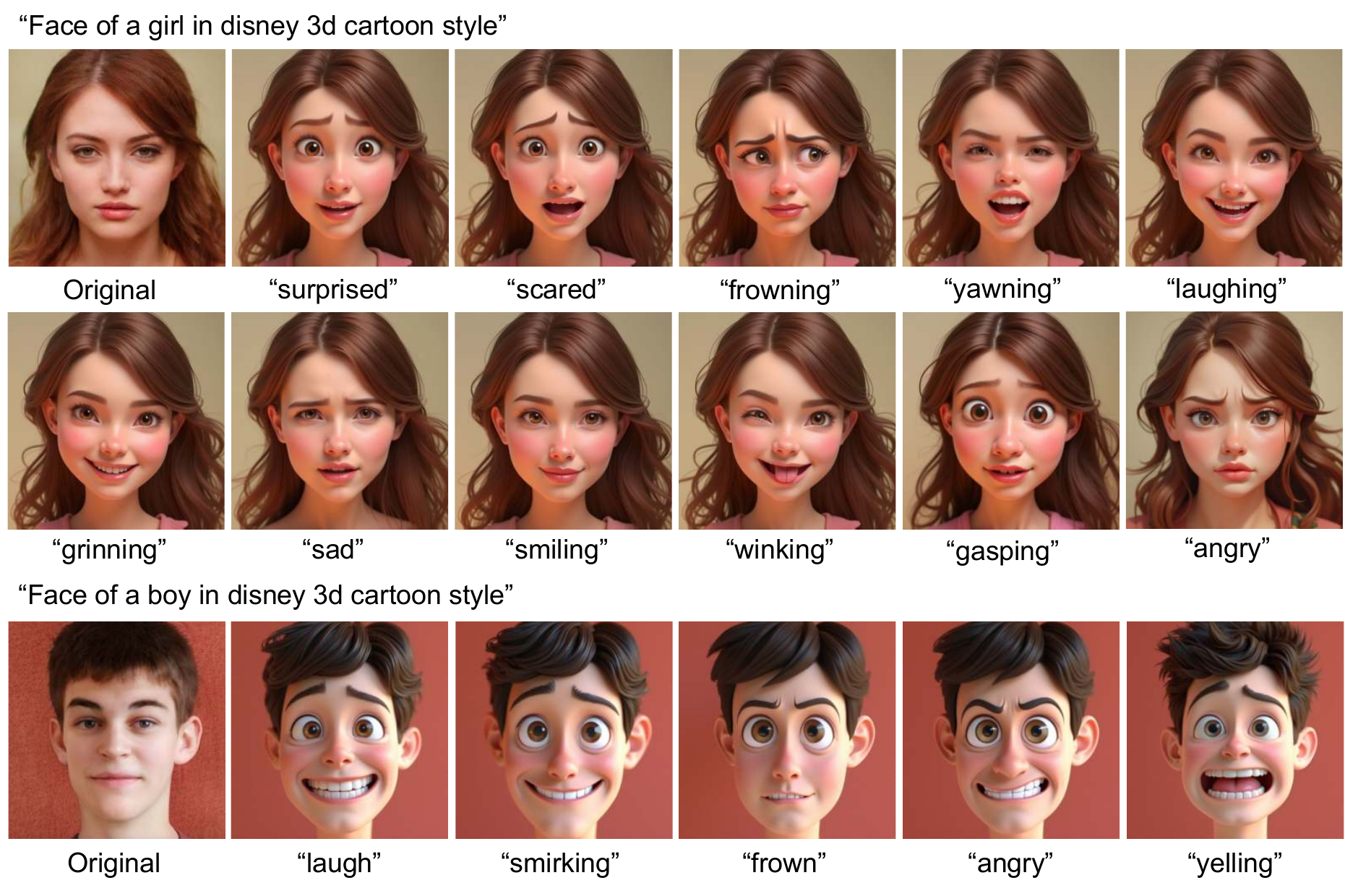}
\caption{
\textbf{Stylization using reference text.}
Stylization of a reference image given prompt-based facial expressions in ``disney 3d cartoon style".
}
\label{fig:expression-edit}
\end{figure}

\Figref{fig:expression-edit} captures a variety of facial expressions that stylize a reference image. 
Given the original image and text prompt: e.g. ``Face of a girl in disney 3d cartoon style", we first invert the image to generate the stylized version of the original image.
Then, we add the prompt for the expression (e.g., ``surprised") at the end of the prompt and run our editing algorithm \eqref{eq:gen-ode-w-controller} with this new prompt: ``Face of a girl in disney 3d cartoon style, surprised". 
By changing the expression, we are able to preserve the identity of the stylized girl and generate prompt-based facial expressions.

\begin{figure}[!tbh]
\includegraphics[width=\linewidth]{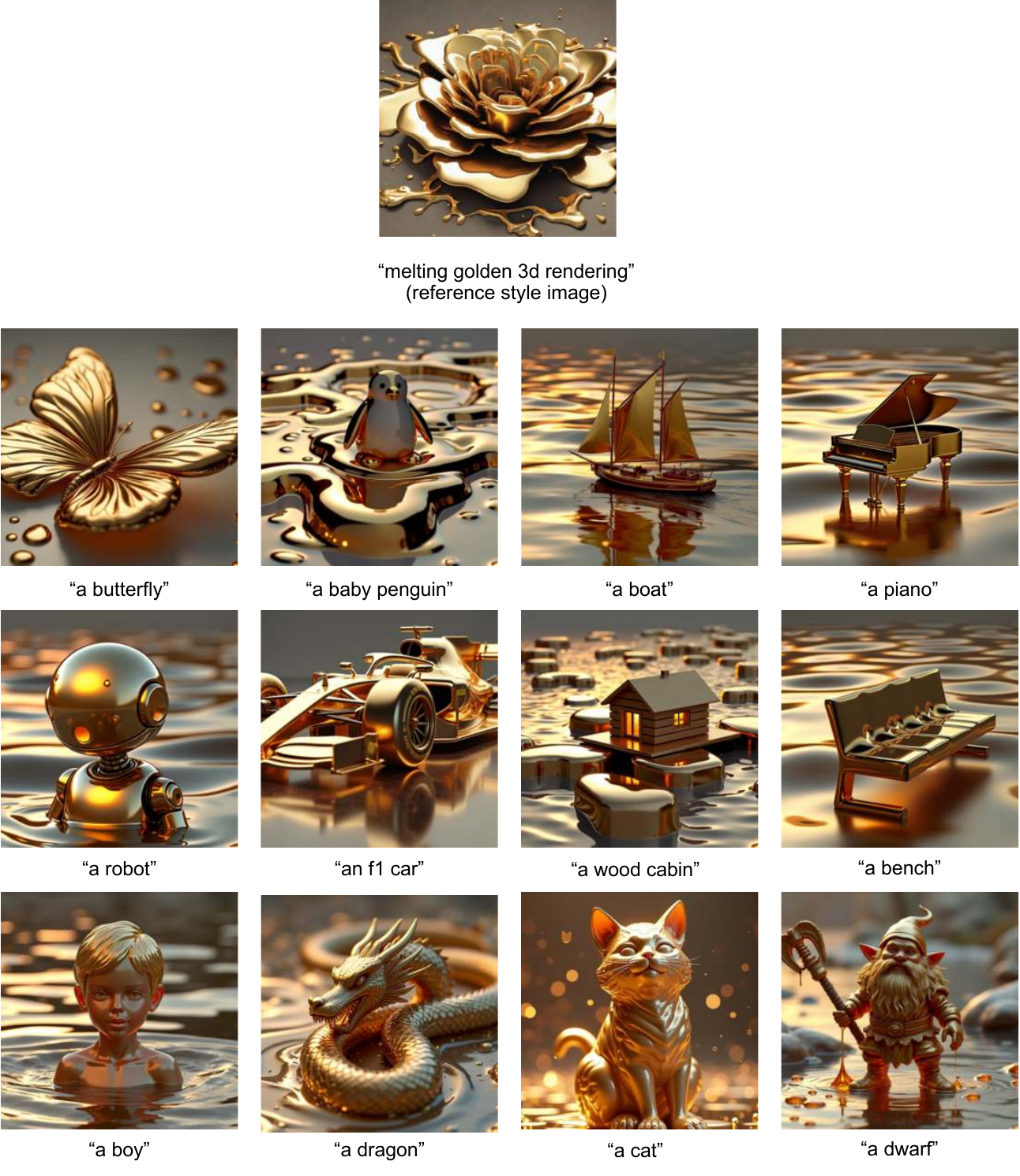}
\caption{
\textbf{Stylization using a single reference image and various text prompts.}
Given a reference style image (e.g. ``melting golden 3d rendering" at the top) and various text prompts (e.g. ``a dwarf in melting golden 3d rendering style"), our method generates images that are consistent with the reference style image and aligned with the given text prompt. 
}
\label{fig:reference-style-all}
\end{figure}

\Figref{fig:reference-style-all} shows stylization based on a single reference style image and 12 different text prompts, covering both living and non-living objects.  
The generated images contain various style attributes that includes melting elements, golden color, and 3d rendering from the reference style image.

\begin{figure}[!tbh]
\includegraphics[width=\linewidth]{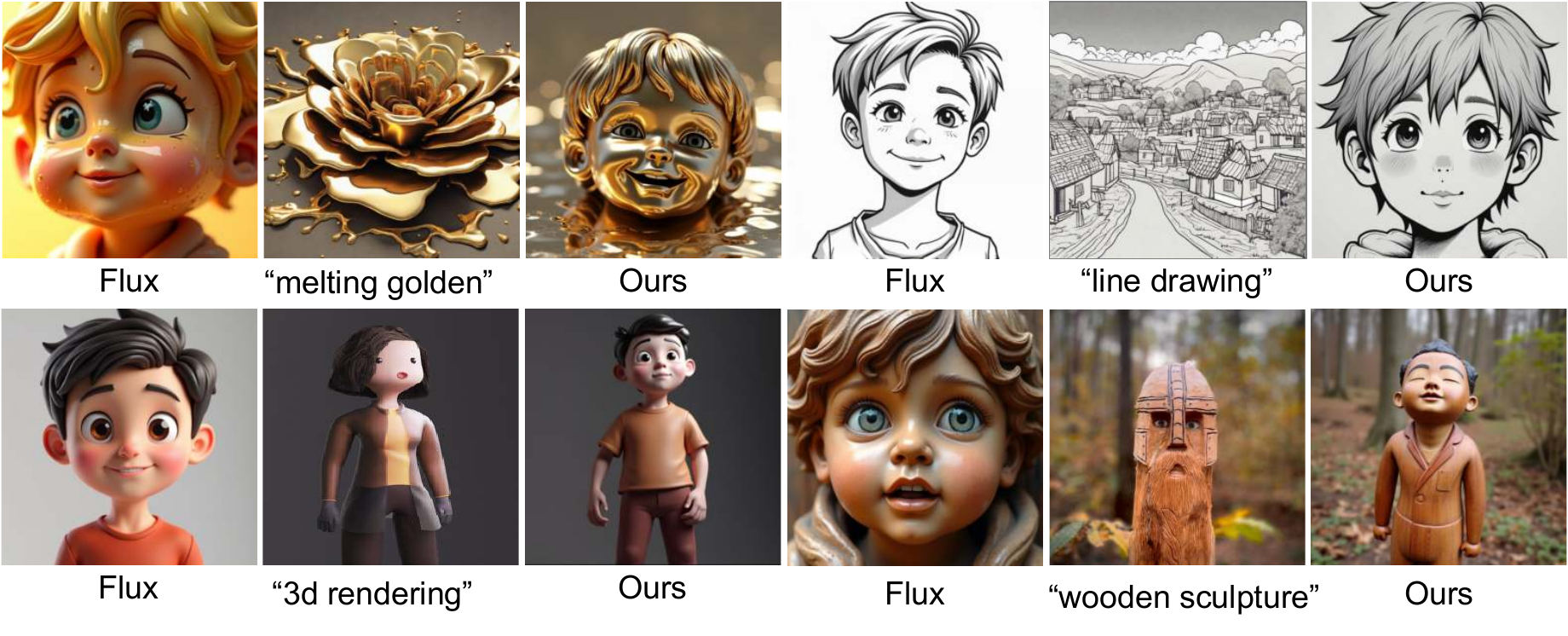}
\caption{
\textbf{Stylization using a single prompt and various reference style images:} ``melting golden", ``line drawing", ``3d rendering", and ``wooden sculpture".
Given a style image (e.g. ``3d rendering") and a text prompt (e.g. ``face of a boy in 3d rendering style"), our method generates images that are consistent with the reference style image and the text prompt. 
The standard output from Flux is obtained by disabling our controller, which clearly highlights the importance of the controller. 
}
\label{fig:reference-style}
\end{figure}

\Figref{fig:reference-style} visualizes stylization results based on different reference style images. In this experiment, we use text prompt to describe both the content of the generated image and the style of the given reference style image.

\begin{figure}[!tbh]
\includegraphics[width=\linewidth]{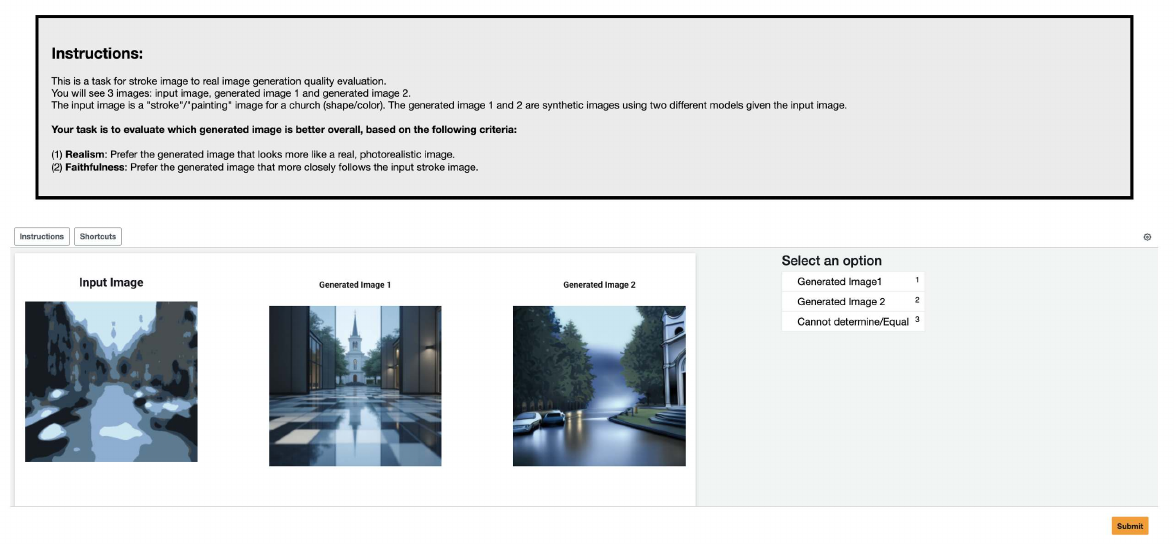}
\caption{
\textbf{Interface for human evaluation.} 
Each participant is asked to select their preferred image based on two criteria: \textit{realism} and \textit{faithfulness}.
}
\label{fig:user-study}
\end{figure}

\subsection{Human Evaluation}
\label{sec:human-eval}
We conduct a user study on the test splits of both LSUN Bedroom and LSUN Church dataset using Amazon Mechanical Turk, with 126 participants in total.
As shown in Figure~\ref{fig:user-study}, each question was accompanied by an explanation of the task, the question, and the evaluation criteria.
Participants were shown a pair of stroke-to-image outputs from different models, in random order, along with the input stroke image. 
They were asked to select one of three options based on their preference using the following two criteria:
\begin{enumerate}
    \item \textbf{Realism:} which of these two images look more like a real, photorealistic image?
    \item \textbf{Faithfulness:} which of these two images match more closely to the input stroke image?
\end{enumerate}

We collect 3 responses per question. 
With 300 images in the test dataset and 10 pairwise comparisons, we gathered 9,000 responses for this evaluation. 
The example in Figure~\ref{fig:user-study} is for the LSUN Church dataset; for LSUN Bedroom dataset, we simply replace the word ``church" to ``bedroom" in the instructions.

\subsection{Generative Modeling using Rectified Stochastic Differential Equations}
\label{sec:appl-rsde}
\begin{figure}[!tbh]
\includegraphics[width=\linewidth]{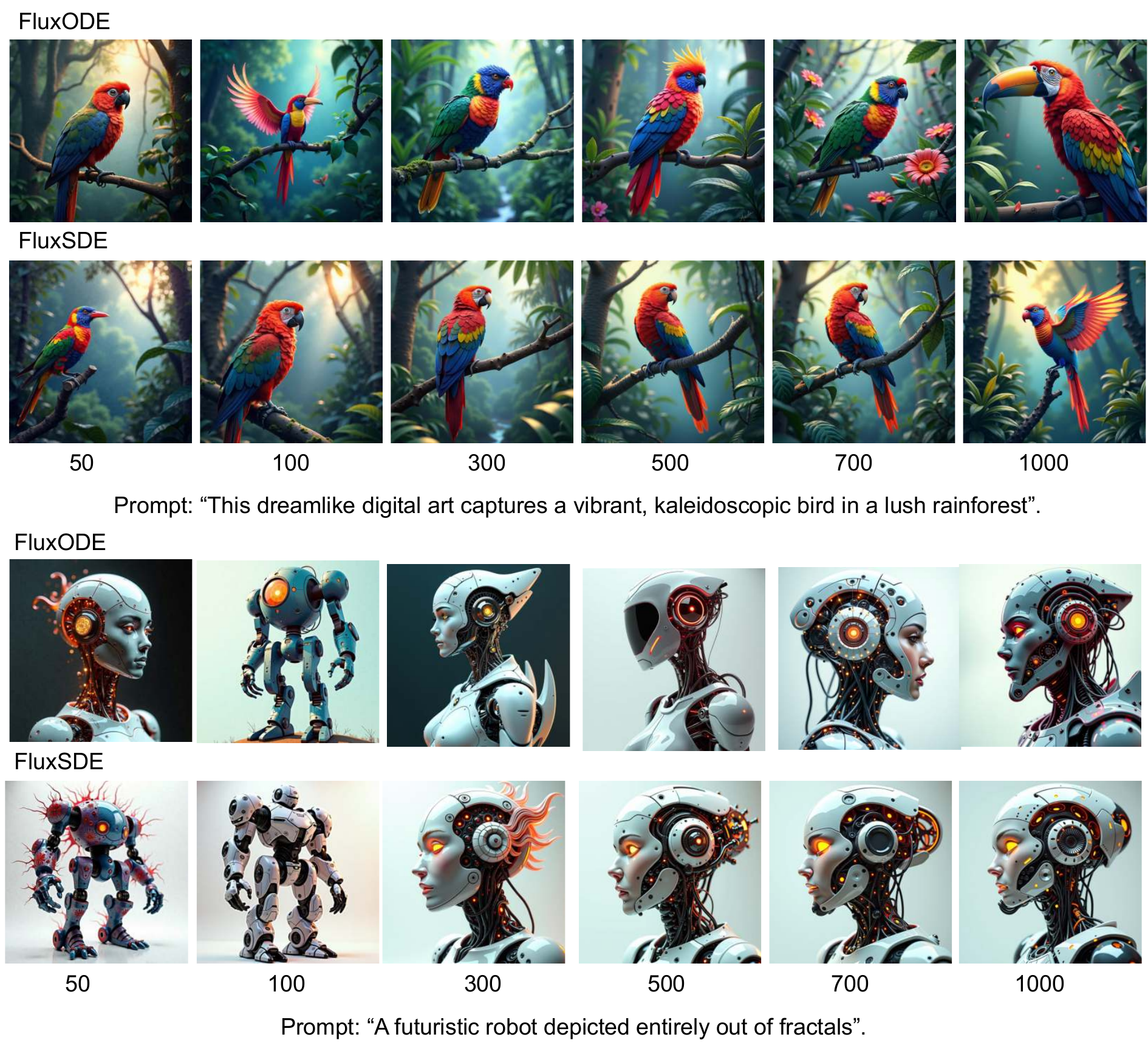}
\caption{\textbf{T2I generation}
using rectified SDE~\eqref{eq:stoch-rect-flow-sampling} for different number of discretization steps marked along the X-axis. The stochastic equivalent sampler FluxSDE generates samples visually comparable to FluxODE at different levels of discretization.
}
\label{fig:appl-resde}
\end{figure}

In \Figref{fig:appl-resde}, we compare images generated by Flux (an ODE-based sampler~\eqref{eq:ode-optimal-vector-field}).
The similarity between the images generated by the ODE and SDE versions of Flux strengthens the practical significance of our theoretical results (\S\ref{sec:method}). 

\end{document}